\documentclass[letterpaper]{article} 
\usepackage{aaai2026}  
\usepackage{times}  
\usepackage{helvet}  
\usepackage{courier}  
\usepackage[hyphens]{url}  
\usepackage{graphicx} 
\usepackage{natbib}  
\usepackage{caption} 
\usepackage{algorithm}
\usepackage{algorithmic}
\usepackage{graphicx} 
\usepackage{multirow}
\usepackage{makecell}
\usepackage{enumitem}
\usepackage{amsmath}

\usepackage{booktabs}
\usepackage{amsmath}
\usepackage{multirow}
\usepackage{bbding}
\usepackage{mathrsfs}
\usepackage{subfigure}
\usepackage{amssymb}
\usepackage{mathtools}
\usepackage{xcolor}
\usepackage[most]{tcolorbox}

\usepackage{pdfpages} %
\usepackage{newfloat}
\usepackage{listings}
\DeclareCaptionStyle{ruled}{labelfont=normalfont,labelsep=colon,strut=off} 
\floatstyle{ruled}
\newfloat{listing}{tb}{lst}{}
\floatname{listing}{Listing}
\title{MCTSr-Zero: Self-Reflective Psychological Counseling Dialogues Generation via Principles and Adaptive Exploration}
\newcommand*\samethanks[1][\value{footnote}]{\footnotemark[#1]}
\author{
    Hao Lu\textsuperscript{\rm 1},
    Yanchi Gu\textsuperscript{\rm 1},
    Haoyuan Huang\textsuperscript{\rm 1},
    Yulin Zhou\textsuperscript{\rm 1},
    Ningxin Zhu\textsuperscript{\rm 1}\thanks{Corresponding author},
    Chen Li\textsuperscript{\rm 1}\samethanks[1]
}

\affiliations{
    \textsuperscript{\rm 1}JianChengXingYun Technology Co., Ltd., Shenzhen, China \\
    \{luhao, zhuningxin, lichen\}@jianchengxingyun.com
}

\begin{document}

\maketitle

\begin{abstract}
The integration of Monte Carlo Tree Search (MCTS) with Large Language Models (LLMs) has demonstrated significant success in structured, problem-oriented tasks. However, applying these methods to open-ended dialogues, such as those in psychological counseling, presents unique challenges. Unlike tasks with objective correctness, success in therapeutic conversations depends on subjective factors like empathetic engagement, ethical adherence, and alignment with human preferences, for which strict ``correctness" criteria are ill-defined. Existing result-oriented MCTS approaches can therefore produce misaligned responses. To address this, we introduce MCTSr-Zero, an MCTS framework designed for open-ended, human-centric dialogues. Its core innovation is ``domain alignment", which shifts the MCTS search objective from predefined end-states towards conversational trajectories that conform to target domain principles (e.g., empathy in counseling). Furthermore, MCTSr-Zero incorporates ``Regeneration" and ``Meta-Prompt Adaptation" mechanisms to substantially broaden exploration by allowing the MCTS to consider fundamentally different initial dialogue strategies. We evaluate MCTSr-Zero in psychological counseling by generating multi-turn dialogue data, which is used to fine-tune an LLM, PsyLLM. We also introduce PsyEval, a benchmark for assessing multi-turn psychological counseling dialogues. Experiments demonstrate that PsyLLM achieves state-of-the-art performance on PsyEval and other relevant metrics, validating MCTSr-Zero's effectiveness in generating high-quality, principle-aligned conversational data for human-centric domains and addressing the LLM challenge of consistently adhering to complex psychological standards.
\end{abstract}

%
\begin{links}
    \link{Code}{https://github.com/JianChengXingYun/Mctsr-Zero} 
\end{links}

\section{Introduction}

The integration of Monte Carlo Tree Search (MCTS) with Large Language Models (LLMs) has recently achieved significant breakthroughs in structured, problem-oriented tasks such as mathematics \citep{zhang2024accessinggpt4levelmathematical,zhang2024llama,guan2025rstar,wang2025mcts,chen2024alphamath,wu2024beyond}. These methods leverage the planning capabilities of MCTS to guide the generative power of LLMs towards optimal solutions.

\begin{figure}[t] 
    \centering
    \includegraphics[width=\columnwidth]{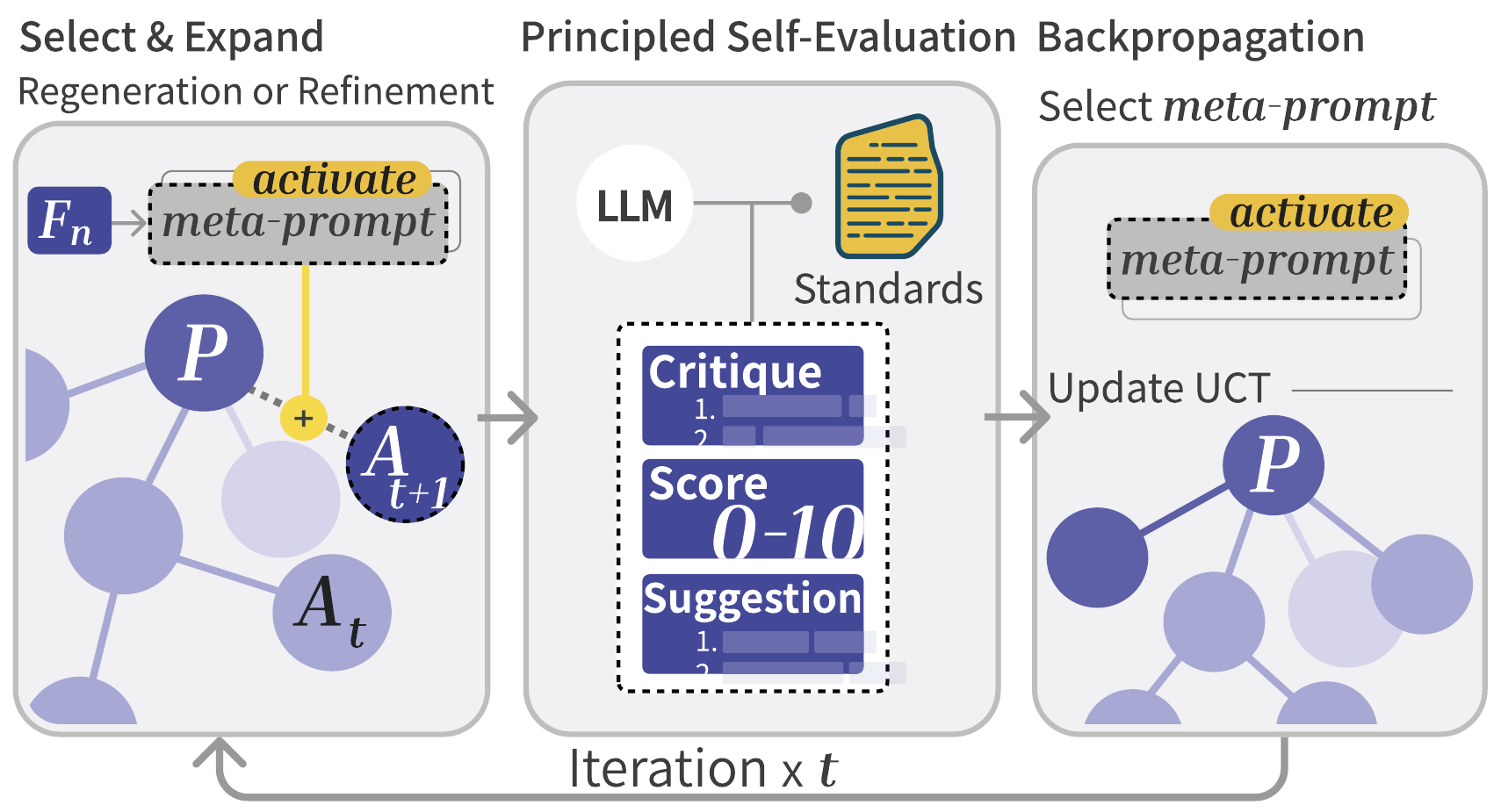}
    \caption{Iterative workflow of MCTSr-Zero: (1) \textbf{Select \& Expand} responses using meta-prompts; (2) \textbf{Principled Self-Evaluation} against standards; (3) \textbf{Backpropagation} updating UCT \& meta-prompts. Iterated $t$ times.}
    \label{fig:figure_1_right_top}
\end{figure}

On another front, recent LLM advancements (e.g., GPT-4 \citep{GPT4.1}) have spurred their application in mental health, leading to specialized models like PsyChat \citep{qiu2024psychat}, CPsyCounX \citep{zhang2024cpsycoun}, Interactive Agents \citep{qiu2024interactive}, and PsyDT \citep{xie2024psydt}. These models often rely on synthesized multi-turn dialogue datasets due to real-data scarcity. A challenge in this domain, however, is that LLMs often struggle to deeply understand and consistently adhere to the complex, abstract, and open-ended psychological standards or principles essential for effective counseling. Given the reliance on synthetic data, enhancing its quality by ensuring better alignment with these principles becomes a key research focus. This points towards the potential of employing search mechanisms—like those offered by MCTS—to actively discover replies more concordant with human preferences and established therapeutic guidelines.

While this potential to leverage MCTS for improved alignment in therapeutic dialogue is promising, applying these methods to open-ended dialogue, especially in domains like psychological counseling, is challenging. Unlike tasks with objective, verifiable answers, success in therapeutic conversations hinges on factors such as empathetic engagement, adherence to ethical guidelines, and alignment with human preferences, for which strict ``correctness" criteria are ill-defined. Consequently, existing result-oriented MCTS-based methods may produce responses misaligned with human expectations or domain-specific conversational goals.

To address the challenge of applying MCTS-enhanced LLMs to open-ended, human-centric dialogues, we introduce MCTSr-Zero. Its core innovation is domain alignment, shifting the search objective from predefined end-states towards conversational trajectories that conform to target domain principles (e.g., empathy in counseling). Furthermore, to broaden MCTS exploration, we propose mechanisms for ``Regeneration" and ``Meta-Prompt Adaptation." These allow the MCTS to explore fundamentally different initial dialogue strategies by modifying the guiding meta-prompt, thereby substantially expanding the search space beyond variations of a single initial approach.

We evaluate MCTSr-Zero in psychological counseling. Specifically, we collected topics from online mental health resources, organized them into case scenarios of psychological counseling, and used MCTSr-Zero to convert these case scenarios into multi-turn dialogue data. This data was used to fine-tune an open-source LLM, named PsyLLM. To address the need for standardized evaluation in this domain, we also developed PsyEval, a benchmark for assessing multi-turn psychological counseling dialogues. Experiments show that PsyLLM, developed using our approach, achieves state-of-the-art (SOTA) performance on PsyEval and other relevant metrics, demonstrating the effectiveness of MCTSr-Zero.

Our contributions are summarized as follows:
\begin{itemize}
\item We propose MCTSr-Zero, an MCTS framework incorporating domain alignment, Regeneration, and Meta-Prompt Adaptation techniques for improved search in open-ended dialogue generation.
\item We construct PsyEval, a benchmark for the automated evaluation of multi-turn dialogues in psychological counseling, addressing a need for standardized assessment.
\item We develop PsyLLM, an LLM for psychological counseling fine-tuned with MCTSr-Zero-generated data, which achieves SOTA performance, demonstrating our approach's effectiveness.
\end{itemize}

\section{Related Work}

\subsection{MCTS-Enhanced LLMs}
Monte Carlo Tree Search (MCTS) has proven highly effective across diverse complex problem domains, from multi-agent pathfinding \citep{pitanov2023monte} and train timetabling \citep{yang2023integrated} to SAT solving \citep{li2023general} and robotics \citep{vagadia2024phyplan}. Recently, MCTS has been integrated with Large Language Models (LLMs) to enhance their reasoning, particularly in structured tasks like mathematics \citep{chen2024alphamath,zhang2024accessinggpt4levelmathematical,zhang2024llama,guan2025rstar,wang2025mcts}, leveraging its planning capabilities to guide LLM generation. However, applying these MCTS-LLM methods to open-ended dialogue, especially in domains like psychological counseling, is challenging because success hinges on subjective factors like empathetic engagement and ethical adherence rather than objectively verifiable correctness, for which strict ``correctness" criteria are ill-defined, potentially leading to responses misaligned with human expectations or domain-specific conversational goals. To address this, our work introduces MCTSr-Zero, which innovatively employs principle-guided domain alignment, shifting the MCTS search objective from predefined end-states towards conversational trajectories that conform to target domain principles.

\subsection{LLMs in Mental Health Support}
Recent LLM advancements (e.g., GPT-4 \citep{GPT4.1}) have spurred their application in mental health Support, leading to specialized models like PsyChat \citep{qiu2024psychat}, CPsyCounX \citep{zhang2024cpsycoun} and Interactive Agents \citep{qiu2024interactive}, which often rely on synthesized multi-turn dialogue datasets due to real-data scarcity. The PsyDT framework \citep{xie2024psydt} further advanced this by personalizing counseling styles. However, a critical challenge is that LLMs often struggle to deeply understand and consistently adhere to the complex, abstract, and open-ended psychological standards or principles essential for effective counseling. Our proposed method employs a search mechanism explicitly guided by these principles to actively discover replies that are more concordant with human preferences and established therapeutic guidelines. This focus on a guided search aims to enhance the LLM's alignment with nuanced therapeutic criteria and thereby improve the intrinsic quality of support provided.

\section{Preliminary}

This section establishes the necessary background on the foundational Monte Carlo Tree Search (MCTS) algorithm, its Upper Confidence Bound (UCB) selection strategy, and the general Monte Carlo Tree Self-Refine (MCTSr) adaptation, which are preliminary to understanding the proposed MCTSr-Zero algorithm detailed in Section \ref{sec:MCTSr-Zero}.

\subsection{Monte Carlo Tree Search (MCTS)}
Monte Carlo Tree Search (MCTS)~\citep{Browne2012ASO} is a heuristic search algorithm widely used in AI for decision-making in complex domains. It builds a search tree iteratively through four phases: \textbf{Selection} traverses the tree using strategies like UCB, \textbf{Expansion} adds new child nodes, \textbf{Simulation} evaluates outcomes, and \textbf{Backpropagation} updates node statistics.

The Upper Confidence Bound (UCB) formula, particularly UCB1, is commonly used in the Selection phase to balance exploration and exploitation. For a node \(a\) and its child \(j\), the UCB value is calculated as:

\begin{equation}
 UCB(a, j) = Q(j) + C \sqrt{\frac{2 \ln N(a)}{N(j)}}
 \label{eq:1}
\end{equation}

where \( Q(j) \) is the estimated value of child node \( j \), \( N(j) \) and \( N(a) \) are visit counts, and \( C \) controls the exploration-exploitation trade-off.

\subsection{Dynamic Monte Carlo Tree Self-Refine (MCTSr)}
Building upon MCTS, the Dynamic Monte Carlo Tree Self-Refine (MCTSr) algorithm adapts this framework for iterative text refinement using LLMs. In MCTSr, nodes represent versions of textual output, and edges represent refinement actions. MCTSr reinterprets the four MCTS phases for text refinement:

\begin{itemize}
    \item \textbf{Selection:} An existing output node is selected for refinement, balancing exploration of different refinement paths (using a strategy like UCB) with exploiting paths that previously yielded high-quality outputs.
    \item \textbf{Expansion:} The selected node is expanded by prompting the LLM with refinement instructions, generating new child nodes with improved text.
    \item \textbf{Evaluation:} Instead of a rollout simulation, MCTSr directly evaluates the quality of a newly generated child node using an LLM-based self-reward function.
    \item \textbf{Backpropagation:} The evaluation outcome (reward/quality) from the evaluated node is propagated up the tree, updating statistics (visit counts and estimated values) for itself and its ancestors.
\end{itemize}
The ``Dynamic" aspect highlights MCTSr's strategies for handling the potentially vast and stochastic nature of LLM-based generation and refinement. Specific notations and the detailed adaptation for psychological consultation dialogue in MCTSr-Zero are presented in Section \ref{sec:MCTSr-Zero}.

\section{MCTSr-Zero: A Framework for Principled, Large-Scale Dialogue Exploration}\label{sec:MCTSr-Zero}

\begin{figure*}[t] 
    \centering
    \includegraphics[width=0.9\textwidth]{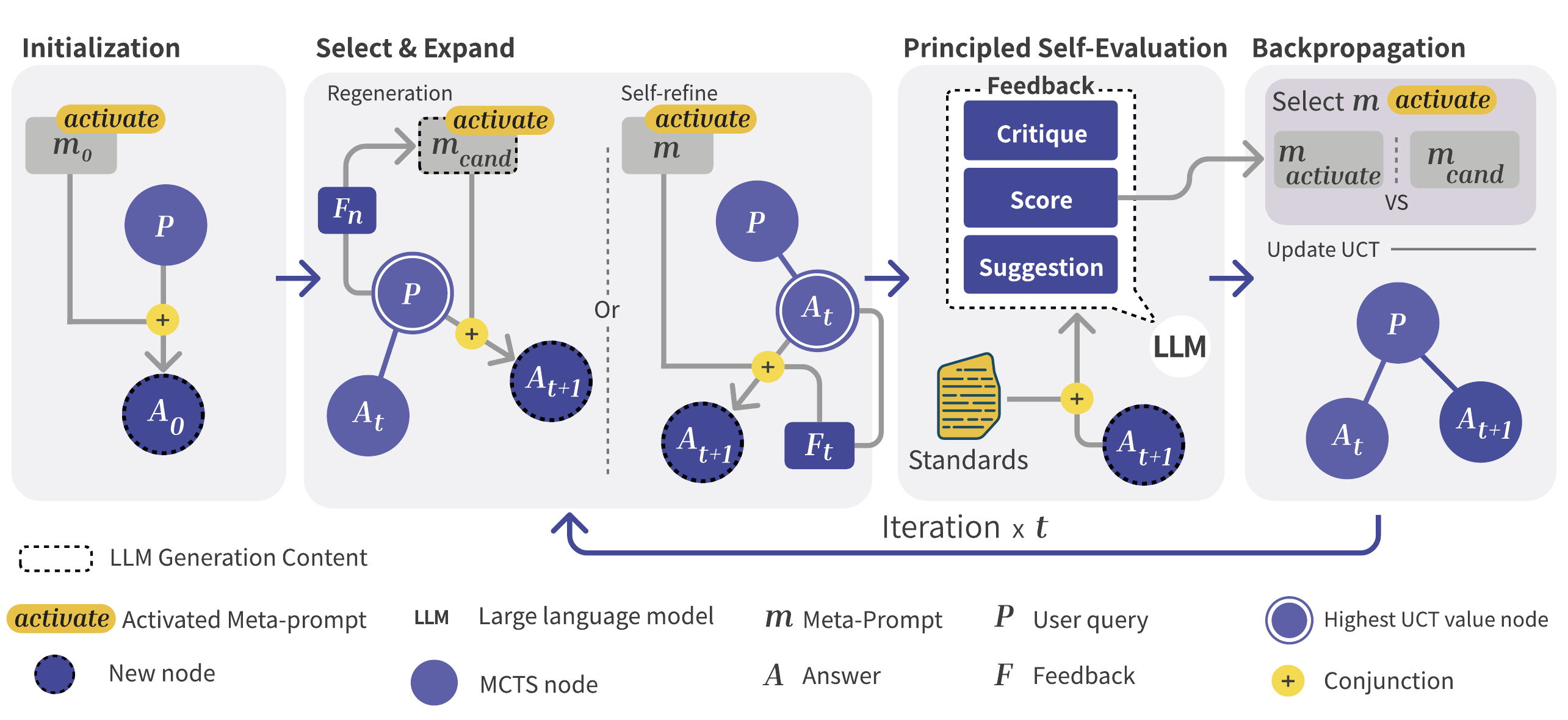}
    \caption{The operational workflow of MCTSr-Zero. (1) \textbf{Initialization}: The meta-prompt $m_0$ is activated to generate initial responses $A_0$ for the user query $P$. (2) \textbf{Select \& Expand}: The system uses the UCT value to either trigger \emph{Regeneration}—using a candidate meta-prompt $m_{\mathrm{cand}}$ as the basis for generating a new answer node $A_{\mathrm{t+1}}$—or trigger \emph{Refinement} to further improve an existing answer node. (3) \textbf{Principled Self-Evaluation}: The new answer is evaluated against predefined standards, producing a critique, score, and actionable suggestions. (4) \textbf{Backpropagation}: Evaluation results are propagated to update UCT scores and guide meta-prompt selection.}
    \label{fig:wide_figure}
\end{figure*}

This section details MCTSr-Zero, an advanced Monte Carlo Tree Search algorithm tailored for generating and refining high-quality, iterative open-ended dialogue. Building upon MCTSr \citep{zhang2024accessinggpt4levelmathematical}, MCTSr-Zero introduces two primary innovations significantly enhancing its capabilities:

\begin{itemize}
    \item Constitutional AI Principles for Self-Alignment and Reflective Iteration: Inspired by Constitutional AI (CAI) \citep{bai2022constitutionalaiharmlessnessai}, MCTSr-Zero empowers the language model with principled self-evaluation and self-refinement. Leveraging predefined psychological standards as a guiding ``constitution," the AI critically assesses its dialogue, identifies areas for improvement, and uses this feedback for iterative refinement. This fosters reflective, self-aligned thinking within the LLM, enhancing dialogue quality and safety without direct human intervention in the core refinement loop.

    \item Vastly Expanded Search Space via Adaptive Exploration: Beyond standard MCTS branch deepening, MCTSr-Zero incorporates mechanisms for exploring a significantly larger, higher-dimensional search space. A UCB-driven selection criterion can choose to refine an existing response or regenerate a new initial response. Coupled with dynamic Meta-Prompt Adaptation, informed by self-evaluation feedback from recently generated initial responses, the algorithm strategically explores fundamentally different initial dialogue strategies. This dual mode—deepening paths via refinement and broadening via new starting points—results in an exponentially larger potential search space than methods limited to variations from a fixed initial strategy.
\end{itemize}

Integrating these innovations allows MCTSr-Zero to not only iteratively improve dialogue quality but also learn and adapt its fundamental generation strategy, guided by explicit principles and empirical self-evaluation. The conceptual structure, illustrating the exploration tree and information flow, is depicted in Figure \ref{fig:wide_figure}.

The main workflow operates in iterative cycles, driving this principled, large-scale exploration through the following stages:
\begin{itemize}
\item Initialization: Setup the root node representing the query and an initial meta-prompt.
\item Selection: Choose a node (either the root or an existing answer node) based on its Upper Confidence Bound for Trees (UCT) value.
\item Action:
\begin{itemize}
\item If the root is selected: A candidate meta-prompt is generated and used to produce a new initial response.
\item If an answer node is selected: That answer undergoes reflective self-refinement.
\end{itemize}
\item Evaluation: The newly generated or refined response is assessed using Principled Self-Evaluation, yielding a score and feedback.
\item Update and Adaptation: The score is backpropagated to update node Q-values. If a new response from a candidate meta-prompt improves the root's value, the active meta-prompt is updated to the candidate. UCT values are then recomputed for the next selection.
\end{itemize}

The algorithm iterates until a termination condition is met.

\subsection{Initialization}
MCTSr-Zero starts with the node $P$ as the root, representing user query $P$. An initial system prompt named meta prompt $m_0$ is set. The language model generates $\mathcal{M}$ initial children $A_0 \dots A_k \sim \mathcal{M}(P\parallel m_0)$, forming $\mathcal{A}_{initial}$. The first generated node, $A_0$, undergoes an immediate evaluation using the Principled Self-Evaluation (Section \ref{sec:principle_eval}). $P$'s initial Q value is then set equal to the reward value obtained from the Principled Self-Evaluation of $A_0$.

\subsection{Selection and Adaptive Exploration Strategy}
In each cycle, UCT (Section \ref{sec:uct_update}) selects node $a \in \mathcal{C}$ (answer nodes $\mathcal{A}$ and $P$), embodying the expanded exploration strategy:

This dynamic UCT choice, $a = \arg\max_{a \in \mathcal{C}} UCT(\mathcal{C})$, balances deepening and broadening the search space.
\begin{itemize}
    \item If $a \in \mathcal{A}$: Self-Refine action on $a$, deepening a path.
    \item If $a = P$: Regeneration action, broadening search via new initial strategy (Meta-Prompt Adaptation).
\end{itemize}

\subsection{Meta-Prompt Adaptation and Search Space Expansion}
A pivotal mechanism for exploring the higher-order search space is Meta-Prompt Adaptation, triggered when $P$ is selected. It allows learning and adjusting the fundamental approach to generating initial responses.

When $P$ is selected at iteration $t+1$, signaling an intent to generate a new initial response, a candidate meta-prompt $m_{cand}$ is synthesized. This synthesis uses the current active meta-prompt $m_{activate}$ and the self-evaluation feedback $\mathcal{F}_{n}$ (critique and suggestions) from a relevant, recently evaluated child node $A_{n} \in \mathcal{A}_{initial}$ of $P$:
\begin{equation}
m_{cand} \leftarrow \mathcal{M}(m_{activate} \parallel \mathcal{F}_{n})
\label{eq:meta_prompt_synthesis}
\end{equation}
This feedback provides targeted insights from recent attempts to generate good initial responses under $P$. Using $m_{cand}$, a new initial response $A_{t+1} = \mathcal{M}(P\parallel m_{cand})$ is generated. This new response $A_{t+1}$ is added to $\mathcal{A}_{initial}$

After $A_{t+1}$ is generated and undergoes Principled Self-Evaluation to obtain its quality score $Q(A_{t+1})$, the active meta-prompt $m_{activate}$ is conditionally updated. If the quality of the new response $Q(A_{t+1})$ is greater than the current quality of the parent node $P$, $Q(P)$ (which represents the average quality of $P$'s children before $A_{t+1}$'s inclusion and subsequent $Q(P)$ update), then $m_{activate}$ is updated to $m_{cand}$:
\begin{equation}
m_{activate} = \begin{cases} m_{cand} & if Q(A_{t+1}) >= Q(P) \\ m_{activate} & otherwise \end{cases}
\label{eq:meta_prompt_update}
\end{equation}
This mechanism allows the system to autonomously learn and switch to better initial generation strategies if they prove more effective than the current average performance of initial responses.

Unlike standard MCTS from a fixed $m_0$, MCTSr-Zero explores from a growing $\mathcal{A}_{initial}$, elevating the search to different distributions $\mathcal{P}(\mathcal{A}_{initial}^{(t)} | P, m_{activate}^{(t)})$. This constitutes exploring a higher-order search space.

\subsection{Reflective Self-Refine}
When a response node $a$ (not $P$) is selected, Reflective Self-Refine transforms $a$ to $a'$. Inspired by Self-Refine \citep{Madaan2023SelfRefineIR}, this uses iterative interaction with the LLM, guided by feedback from $a$'s Principled Self-Evaluation. The model uses concrete, standards-based critique and suggestions $\mathcal{F}$ and meta prompt $m_{activate}$ as explicit guidance to produce $A_{t+1}' = \mathcal{M}(A_{t} \parallel \mathcal{F}_{t} \parallel m_{activate})$. This enables targeted improvement based on AI's principled assessment and Self-Refine techniques.

\subsection{Principled Self-Evaluation: Applying the Constitution}\label{sec:principle_eval}
Principled Self-Evaluation is the cornerstone, providing crucial feedback for refinement and adaptation. It rigorously assesses new/refined response $a$. Inspired by Constitutional AI \citep{bai2022constitutionalaiharmlessnessai}, which uses principles for self-improvement, we use 16 psychological standards as our AI's ``constitution."

For response $a$, the LLM performs structured evaluation guided by standards:
\begin{itemize}
    \item Critique based on Constitution: Analyzes $a$ against 16 standards, identifying strengths/weaknesses relative to these principles. This critique guides subsequent steps towards standard fulfillment.
    \item Scoring (0-10): Assigns a reward score based on critique and adherence to standards.
    \item Actionable Suggestions: Provides specific suggestions for improving $a$ to better meet standards. These are used as guidance for Refinement and the feedback for Meta-Prompt Adaptation as described in Eq. \ref{eq:meta_prompt_synthesis}.
\end{itemize}

This embodies Constitutional AI: AI uses principles to evaluate output and generate feedback for self-improvement, fostering self-alignment. Evaluation adheres strictly to standards and is sampled multiple times for robustness. $Q(a)$ from sampled scores $R_a$ uses:
\begin{equation}
Q(a) = \frac{1}{2}\left(\min{R_a} + \frac{1}{|R_a|}\Sigma^{|R_a|}_{i=1}{R_a^i}\right)
 \label{eq:2_updated_psy}
\end{equation}
This balances average quality and low-score robustness.

\subsection{Backpropagation}
Following Principled Self-Evaluation, Q value and insights propagate upwards.
\begin{itemize}
    \item For evaluated response $a$, $Q(a)$ is set using Eq. \ref{eq:2_updated_psy}.
    \item For parent answer node $p$ of $a$, $Q^{\prime}(p)$ updates recursively:
    \begin{equation}
    Q^{\prime}(p) = \frac{1}{2} \left(Q(p) + \max_{c \in Children} Q(c)\right)
     \label{eq:3_updated_psy}
    \end{equation}
    \item For user query node $P$, $Q(P)$ is the average of its children's Q values:
    \begin{equation}
    Q(P) = \frac{1}{|\mathcal{A}_{initial}|} \sum_{a \in \mathcal{A}_{initial}} Q(a)
     \label{eq:query_q_psy}
    \end{equation}
    This $Q(P)$ value is used in the condition for updating the active meta-prompt (Eq. \ref{eq:meta_prompt_update}).
\end{itemize}
Backpropagation ensures performance signals inform future decisions.

\subsection{Update UCT and Guiding Large-Scale Exploration}\label{sec:uct_update}
After Backpropagation, UCT values recompute for selectable nodes $\mathcal{C}$ answer nodes and $P$). UCT is computed using UCB:
\begin{equation}
UCT_s = Q(s) + c \sqrt{\frac{\ln{N(Parent(s)) + 1}}{N(s) + \epsilon}}
 \label{eq:4_updated_psy}
\end{equation}
where $Q(s)$ is quality (Eq. \ref{eq:2_updated_psy} or Eq. \ref{eq:query_q_psy}), $N(s)$ is visit count, $N(Parent(s))$ is parent visit count (adjusted for $P$), $c$ balances exploration/exploitation, $\epsilon$ prevents division by zero. $a \in \mathcal{C}$ with highest UCT is selected. If $a$ is answer node, high UCT signals refining this path. If $a$ is $P$, high UCT signals exploring a new initial strategy via Regeneration/Meta-Prompt Adaptation. This UCT-driven selection intelligently navigates the massively expanded search space, balancing deepening paths with exploring new starting points, guided by principled self-evaluation feedback.

\subsection{Termination Function}
Iteration continues until condition $T$ is met (e.g., budget, depth, quality threshold, stagnation). Final output is the dialogue response node with the highest Q value.


\section{PsyEval Benchmark: Evaluating AI Empathy in Psychological Support}\label{sec:psyeval}
The growing use of AI for psychological support demands evaluation beyond standard metrics. Conventional benchmarks often fail to assess crucial competencies like accurate empathy, re-framing, and meaningful questioning. To bridge this gap, we introduce PsyEval—a specialized, multi-dimensional benchmark designed to gauge AI competence in simulated counseling dialogues grounded in realistic scenarios.

\subsection{Systematic Scenario and Data Generation}
PsyEval creates diverse psychological scenarios by synthesizing detailed case reports, rather than relying on existing human data. Based on these reports, a Language Model (LLM) simulates a multi-turn counseling dialogue. An independent AI Judge then evaluates this dialogue against 16 predefined criteria. This systematic process enables rigorous testing of AI capabilities across various psychological contexts. Details on scenarios and criteria are in Appendix A and B, which are provided in the supplementary material.

\subsection{Comprehensive Multi-Dimensional Evaluation}\label{sec:multi-eval}
The core of PsyEval is its novel 16-dimension evaluation framework, which assesses an AI therapist's ability to provide empathic support in multi-turn interactions. Adopting a third-party observational perspective similar to the ESHCC benchmark \citep{concannon2024measuring}, it focuses on an AI's observable expressions of counseling skill. The rubric uses inferential language (e.g., “the system seems to…”) to capture a neutral observer's perception of the interaction quality.
The 16 dimensions synthesize insights from established frameworks, including the Therapist Empathy Scale (TES) \citep{decker2014development}, ESHCC, Motivational Interviewing (MI) \citep{bolton2021theoretical}, and Person-centered therapy \citep{rogers2007counseling}. We optimized existing dimensions and integrated six new ones crucial for psychological counseling: Dialogical Logical Consistency, Conversational Continuity, Resistance Handling, Ethics/Prosocial Guidance, Summarizing, and Dialogue Pacing/Process Attunement. We also redefine the “Fallacy Avoidance” dimension to evaluate hallucination control—the AI’s ability to remain coherent and factually grounded while maintaining a consistent persona.
Unlike clinical-only scales like TES, our benchmark has broader applicability, evaluating AI in both therapeutic and everyday emotional support contexts. This “dual role" enhances its ecological validity, assessing AI as both a structured therapist and an empathetic companion.

\subsection{AI-based Judging Mechanism}
PsyEval employs an independent AI Judge for evaluation, configured with our 16-dimension rubric. This approach ensures scalable, efficient, and consistent scoring across numerous dialogues, minimizing the human rater variability common in subjective assessments. Details of the AI judge setup are in Appendix C.
These standards, developed by human experts and tailored for LLM comprehension, are designed to ensure high alignment between the AI's evaluation and human judgment, thereby minimizing potential evaluation bias.

\section{Experiment}


\begin{figure}[t] 
    \centering
    \includegraphics[width=\columnwidth]{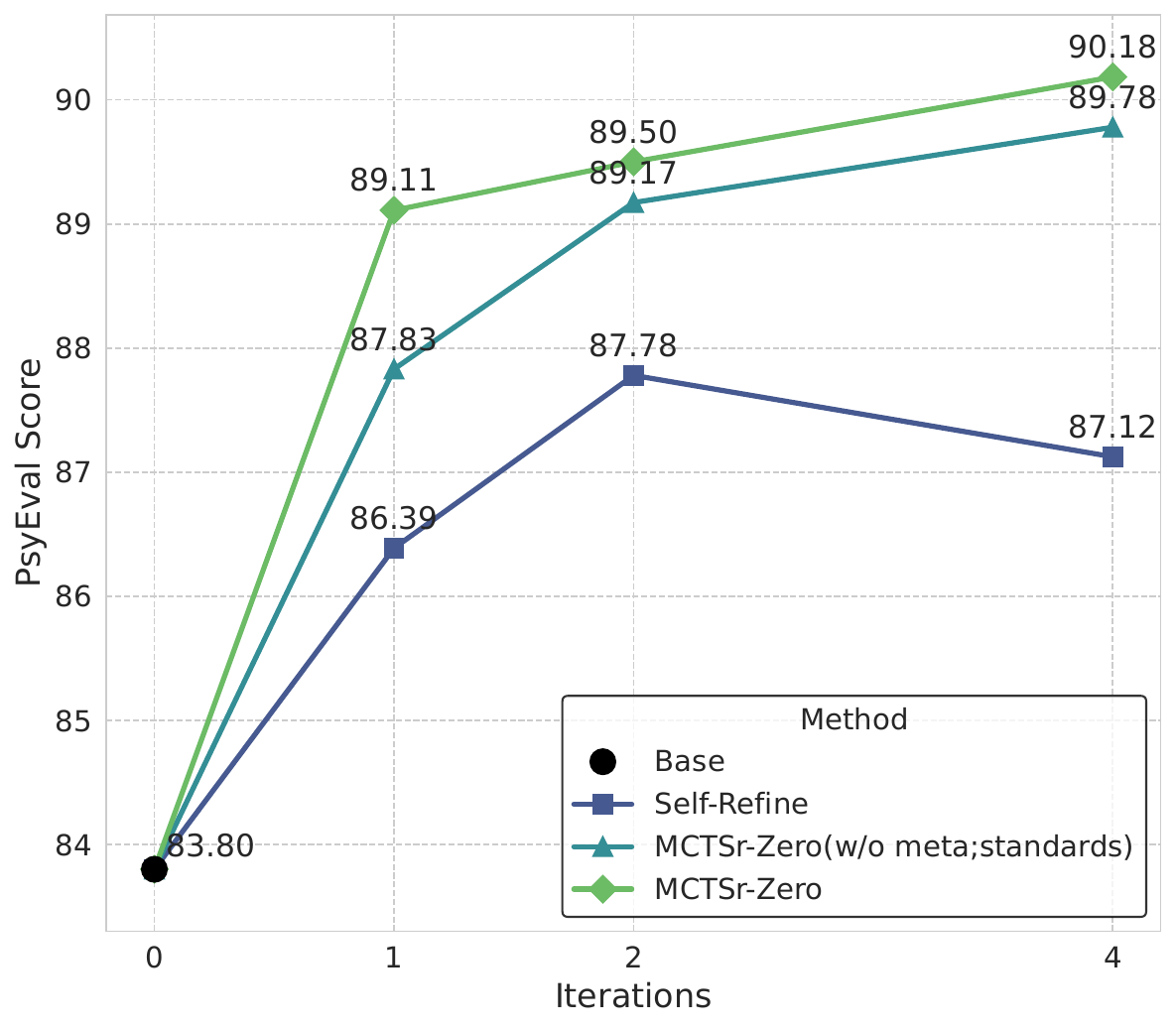}
    \caption{PsyEval scores for gpt-4.1-mini with four different refinement methods across iterations. MCTSr-Zero demonstrates the highest performance.}
    \label{fig:figure_3_left_top}
\end{figure}

\begin{table*}[!t]
    \centering
    \setlength{\tabcolsep}{2mm} 
    \begin{tabular}{llcccccccc} 
    \toprule
    {\textbf{Type}} & {\textbf{Model Name}} & {\makecell[c]{\textbf{Total}\\\textbf{Score}}} & {\makecell[c]{\textbf{ESHCC}\\\textbf{REVISED}}} & {\makecell[c]{\textbf{DLC}}} & {\makecell[c]{\textbf{CC}}} & {\makecell[c]{\textbf{RH}}} & {\makecell[c]{\textbf{Sum.}}} & {\makecell[c]{\textbf{EPG}}} & {\makecell[c]{\textbf{DPPA}}} \\
    \midrule
    \multirow{2}{*}{Ours} & \textbf{PsyLLM-Large-250519} & \textbf{90.93} & \textbf{54.53} & \textbf{9.16} & 4.57 & 4.56 & \textbf{4.47} & 4.53 & \textbf{4.55} \\
    & PsyLLM-Mini-250519 & 90.72 & 54.46 & 9.15 & \textbf{4.58} & \textbf{4.57} & 4.43 & 4.47 & 4.51 \\
    \midrule
    \multirow{16}{*}{Other} & claude-3-7-sonnet-20250219 & 88.89 & 53.13 & 9.03 & 4.51 & 4.44 & 4.28 & \textbf{4.56} & 4.49 \\
    & gemini-2.5-pro-exp-03-25 & 88.62 & 53.01 & 9.06 & 4.53 & 4.48 & 4.33 & 4.34 & 4.36 \\
    & gemini-2.5-flash-preview-04-17 & 88.07 & 52.59 & 8.94 & 4.50 & 4.47 & 4.27 & 4.39 & 4.39 \\
    & gpt-4.1 & 85.65 & 50.87 & 8.77 & 4.44 & 4.44 & 4.04 & 4.32 & 4.38 \\
    & gpt-4.1-mini & 83.80 & 49.82 & 8.69 & 4.38 & 4.19 & 3.94 & 4.18 & 4.21 \\
    & gpt-4o-2024-11-20 & 82.31 & 48.71 & 8.52 & 4.28 & 4.18 & 3.87 & 4.25 & 4.24 \\
    & Qwen3-32B-w/o Reasoning & 81.40 & 48.66 & 8.45 & 4.28 & 4.16 & 3.78 & 4.06 & 4.10 \\
    & GLM-4-32B-0414 & 80.92 & 48.00 & 8.39 & 4.25 & 3.97 & 3.82 & 4.05 & 4.13 \\
    & gpt-4.1-nano & 80.72 & 47.58 & 8.52 & 4.25 & 3.96 & 3.83 & 4.14 & 4.10 \\
    & qwen-max-2025-01-25 & 79.59 & 47.11 & 8.32 & 4.22 & 3.99 & 3.65 & 4.14 & 3.93 \\
    & gpt-4o-mini-2024-07-18 & 78.76 & 46.48 & 8.21 & 4.25 & 3.82 & 3.59 & 4.06 & 4.10 \\
    & doubao-1-5-pro-32k-250115 & 78.71 & 46.80 & 8.19 & 4.17 & 3.89 & 3.66 & 3.97 & 3.90 \\
    & Qwen2.5-72B-Instruct & 76.41 & 45.53 & 7.97 & 3.98 & 3.61 & 3.48 & 3.92 & 3.82 \\
    & GLM-4-9B-0414 & 75.74 & 44.99 & 7.94 & 3.96 & 3.76 & 3.41 & 3.91 & 3.79 \\
    & Qwen2.5-32B-Instruct & 75.70 & 44.54 & 8.08 & 4.02 & 3.67 & 3.47 & 4.03 & 3.91 \\
    & doubao-1-5-lite-32k-250115 & 74.99 & 44.72 & 7.88 & 4.00 & 3.55 & 3.32 & 3.91 & 3.59 \\
    \midrule
    \multirow{4}{*}{Domain} & simpsybot\_D & 77.92 & 45.87 & 8.36 & 4.15 & 3.74 & 3.60 & 4.13 & 3.92 \\
    & SoulChat2\_0-Qwen2-7B & 77.38 & 45.55 & 8.08 & 4.20 & 3.84 & 3.56 & 4.10 & 3.98 \\
    & Xinyuan-LLM-14B-0428 & 76.41 & 45.26 & 8.02 & 4.02 & 3.72 & 3.50 & 4.07 & 3.78 \\
    & CPsyCounX & 66.00 & 39.99 & 6.76 & 3.37 & 3.24 & 3.01 & 3.82 & 3.31 \\
    \bottomrule
    \end{tabular}
    \caption{Results on the PsyEval Benchmark. Scores for each criterion reflect model performance in simulated psychological counseling dialogues. The criteria are abbreviated as follows: ESHCC-R (Revised Empathic Support, Human Connection, and Care), DLC (Dialogical Logical Consistency), CC (Conversational Continuity), RH (Resistance Handling), Sum. (Summarizing), EPG (Ethics and Prosocial Guidance), and DPPA (Dialogue Pacing, Process, and Attunement). The best result in each column is highlighted in \textbf{bold}.}
    \label{tab:psyeval_results}
\end{table*}

This section presents the experimental evaluation of various language models on the PsyEval Benchmark. We report results from two key experiments: (1) a comprehensive benchmark comparison of our PsyLLM variants against leading baseline models, including other psychological domain-specific models, evaluated using PsyEval, and (2) an ablation study demonstrating the impact of iterative refinement methods aligned with principles of frameworks like MCTSr-Zero.

\subsection{MCTSr-Zero-Psy}
To validate the effectiveness of our proposed dialogue reconstruction method, we generated dialogues using MCTSr-Zero. We denote this collection as MCTSr-Zero-Psy, which comprises 4,000 multi-turn counseling dialogues. These dialogues cover 16 distinct categories of psychological distress (as detailed in Section \ref{sec:multi-eval}), with each category containing approximately 250 entries, and an average of 20 turns per dialogue.

\subsection{PsyLLM}
Our PsyLLM model, developed in Large (based on GLM-4-32B-0414) and Mini (based on GLM-4-9B-0414) variants, underwent a two-stage training process on 4 NVIDIA A800 GPUs. We used MCTSr-Zero-Psy as our training data. First, for supervised fine-tuning (SFT), each respective pre-trained GLM-4 base model was trained for 2 epochs with a learning rate of $1 \times 10^{-4}$, a 0.1 linear warmup, a per-device batch size of 1, and the AdamW \citep{loshchilov2019decoupledweightdecayregularization} optimizer. Subsequently, the SFT-trained models underwent SimPO \citep{NEURIPS2024_e099c1c9} alignment for 3 epochs, employing a reduced learning rate of $5 \times 10^{-7}$, maintaining a 0.1 linear warmup and a per-device batch size of 1 with the AdamW optimizer, ultimately yielding the final PsyLLM models.

\subsection{Experimental Setup}
Our evaluation utilizes the PsyEval Benchmark, which simulates psychological counseling dialogues for 64 unique case scenarios. An AI Judge evaluates responses against \textbf{16} specific psychological consultation criteria to calculate a total score. Further details on the AI Judge are in Appendix C.

We conduct two main experiments:

\begin{itemize}
    \item \textbf{Main Benchmark Comparison on PsyEval:} We comprehensively evaluate a wide array of models, including leading commercial, general-purpose open-source, and other psychological domain-specific models. This includes our PsyLLM-Large-250519 (based on GLM-4-32B-0414\cite{glm2024chatglm}) and PsyLLM-Mini-250519 (based on GLM-4-9B-0414) models, which were developed using techniques aligned with MCTSr-Zero's principles of iterative refinement. Results are presented in Table \ref{tab:psyeval_results}.
    
    \item \textbf{Ablation Study on Iterative Refinement Methods:} We investigate the effectiveness of iterative refinement on a \texttt{gpt-4.1-mini} base model. We apply \texttt{Self-Refine}, \texttt{MCTSr-Zero(w/o meta;standards)} (a variant using \texttt{Self-Refine} for node refinement instead of standard-based evaluation), and the full \texttt{MCTSr-Zero} framework. These methods are tested across 1, 2, and 4 iterations and compared to a non-refined baseline (0 iterations), with results shown in Figure \ref{fig:figure_3_left_top}.
\end{itemize}

\subsection{Analysis of Benchmark Results}

Table \ref{tab:psyeval_results} details the performance of all models on the PsyEval Benchmark, as assessed by the AI Judge across all 16 criteria. The data indicates that our proposed models, PsyLLM-Large and PsyLLM-Mini, achieved the highest overall scores of 90.93 and 90.72, respectively. This demonstrates a discernible advantage over other evaluated models, including the next-highest score of 88.89.

A deeper analysis reveals that this strong performance is not limited to a single capability but reflects a balanced and holistic profile. For instance, the PsyLLM variants achieved high scores in empathetic and human-centered communication (consolidated as `ESHCC REVISED` in the table for brevity) while also excelling in maintaining logical consistency, conversational continuity, and effectively handling user resistance.

We attribute this robust and well-rounded performance to the direct alignment between our development methodology and the evaluation benchmark. The performance disparity seen in the results stems from a core design principle: the PsyEval criteria are derived from the same 16 psychological standards that guide the iterative refinement within our proposed framework, MCTSr-Zero. The high scores of the PsyLLM variants underscore the value of a synergistic approach that combines domain-specific evaluation (PsyEval) with a tailored development framework (MCTSr-Zero). This deep alignment is fundamental to building effective and responsible AI for psychological support.

\subsection{Ablation Study}
Figure \ref{fig:figure_3_left_top} shows our ablation study results, demonstrating that iterative refinement methods substantially improve the performance of a base model (gpt-4.1-mini) on PsyEval.

The baseline model's score of 83.60 improves significantly with just one iteration of Self-Refine (to 86.39) or an MCTSr-Zero variant (over 87). Performance generally increases with more iterations. The full \texttt{MCTSr-Zero} framework consistently outperforms simpler methods and its ablated variant, reaching a peak score of 90.18 after 4 iterations. This study validates that iterative processes and strategic search—core mechanisms of frameworks like MCTSr-Zero—effectively enhance an AI's capabilities for psychological support by leveraging self-reflection and evaluation against explicit standards.

\section{Conclusion}
We introduced MCTSr-Zero, an advanced Monte Carlo Tree Search algorithm that uses dynamic meta-prompts and principles-based self-evaluation to generate high-quality dialogues for training PsyLLM, our specialized language model for psychological counseling.

To assess such models, we also developed the PsyEval Benchmark, which uses an AI Judge to evaluate dialogues against 16 psychological criteria. Our results show that PsyLLM models, developed using MCTSr-Zero principles, significantly outperform leading general and domain-specific models. This underscores the value of our specialized framework and benchmark for the nuanced demands of psychological consultation.

Key limitations include MCTSr-Zero's computational cost and potential biases in PsyEval's AI Judge. Future work will focus on improving MCTSr-Zero's search efficiency and refinement techniques, while simultaneously advancing PsyEval by mitigating bias, expanding its scenarios, and integrating human evaluation.



\nocite{*} 
\bibliography{aaai26}

@misc{zhang2024accessinggpt4levelmathematical,
      title={Accessing GPT-4 level Mathematical Olympiad Solutions via Monte Carlo Tree Self-refine with LLaMa-3 8B}, 
      author={Di Zhang and Xiaoshui Huang and Dongzhan Zhou and Yuqiang Li and Wanli Ouyang},
      year={2024},
      eprint={2406.07394},
      archivePrefix={arXiv},
      primaryClass={cs.AI},
      url={https://arxiv.org/abs/2406.07394}, 
}

@misc{guan2025rstar,
      title={rStar-Math: Small LLMs Can Master Math Reasoning with Self-Evolved Deep Thinking}, 
      author={Xinyu Guan and Li Lyna Zhang and Yifei Liu and Ning Shang and Youran Sun and Yi Zhu and Fan Yang and Mao Yang},
      year={2025},
      eprint={2501.04519},
      archivePrefix={arXiv},
      primaryClass={cs.CL},
      url={https://arxiv.org/abs/2501.04519}, 
}

@misc{wang2025mcts,
      title={MCTS-Judge: Test-Time Scaling in LLM-as-a-Judge for Code Correctness Evaluation}, 
      author={Yutong Wang and Pengliang Ji and Chaoqun Yang and Kaixin Li and Ming Hu and Jiaoyang Li and Guillaume Sartoretti},
      year={2025},
      eprint={2502.12468},
      archivePrefix={arXiv},
      primaryClass={cs.LG},
      url={https://arxiv.org/abs/2502.12468}, 
}

@misc{zhang2024llama,
      title={LLaMA-Berry: Pairwise Optimization for O1-like Olympiad-Level Mathematical Reasoning}, 
      author={Di Zhang and Jianbo Wu and Jingdi Lei and Tong Che and Jiatong Li and Tong Xie and Xiaoshui Huang and Shufei Zhang and Marco Pavone and Yuqiang Li and Wanli Ouyang and Dongzhan Zhou},
      year={2024},
      eprint={2410.02884},
      archivePrefix={arXiv},
      primaryClass={cs.AI},
      url={https://arxiv.org/abs/2410.02884}, 
}

@misc{chen2024alphamath,
      title={AlphaMath Almost Zero: Process Supervision without Process}, 
      author={Guoxin Chen and Minpeng Liao and Chengxi Li and Kai Fan},
      year={2024},
      eprint={2405.03553},
      archivePrefix={arXiv},
      primaryClass={cs.CL},
      url={https://arxiv.org/abs/2405.03553}, 
}

@misc{wu2024beyond,
      title={Beyond Examples: High-level Automated Reasoning Paradigm in In-Context Learning via MCTS}, 
      author={Jinyang Wu and Mingkuan Feng and Shuai Zhang and Feihu Che and Zengqi Wen and Jianhua Tao},
      year={2024},
      eprint={2411.18478},
      archivePrefix={arXiv},
      primaryClass={cs.CL},
      url={https://arxiv.org/abs/2411.18478}, 
}

@inproceedings{pitanov2023monte,
  title={Monte-carlo tree search for multi-agent pathfinding: Preliminary results},
  author={Pitanov, Yelisey and Skrynnik, Alexey and Andreychuk, Anton and Yakovlev, Konstantin and Panov, Aleksandr},
  booktitle={International Conference on Hybrid Artificial Intelligence Systems},
  pages={649--660},
  year={2023},
  organization={Springer}
}

@misc{yang2023integrated,
      title={An Integrated Framework Integrating Monte Carlo Tree Search and Supervised Learning for Train Timetabling Problem}, 
      author={Feiyu Yang},
      year={2023},
      eprint={2311.00971},
      archivePrefix={arXiv},
      primaryClass={cs.LG},
      url={https://arxiv.org/abs/2311.00971}, 
}

@misc{li2023general,
      title={General Method for Solving Four Types of SAT Problems}, 
      author={Anqi Li and Congying Han and Tiande Guo and Haoran Li and Bonan Li},
      year={2023},
      eprint={2312.16423},
      archivePrefix={arXiv},
      primaryClass={cs.AI},
      url={https://arxiv.org/abs/2312.16423}, 
}

@misc{vagadia2024phyplan,
      title={PhyPlan: Compositional and Adaptive Physical Task Reasoning with Physics-Informed Skill Networks for Robot Manipulators}, 
      author={Harshil Vagadia and Mudit Chopra and Abhinav Barnawal and Tamajit Banerjee and Shreshth Tuli and Souvik Chakraborty and Rohan Paul},
      year={2024},
      eprint={2402.15767},
      archivePrefix={arXiv},
      primaryClass={cs.RO},
      url={https://arxiv.org/abs/2402.15767}, 
}

@misc{xu2023no,
      title={No Train Still Gain. Unleash Mathematical Reasoning of Large Language Models with Monte Carlo Tree Search Guided by Energy Function}, 
      author={Haotian Xu},
      year={2023},
      eprint={2309.03224},
      archivePrefix={arXiv},
      primaryClass={cs.AI},
      url={https://arxiv.org/abs/2309.03224}, 
}

@misc{GPT4.1,
year={2025},
author={{OpenAI}},
	title = {Introducing GPT-4.1 in the API},
	howpublished = "\url{https://openai.com/index/gpt-4-1/}",
	note = "Accessed: 2025-04-14",
}

@misc{achiam2023gpt,
      title={GPT-4 Technical Report}, 
      author={Achiam, Josh and Adler, Steven and Agarwal, Sandhini and Ahmad, Lama and Akkaya, Ilge and Aleman, Florencia Leoni and Almeida, Diogo and Altenschmidt, Janko and Altman, Sam and Anadkat, Shyamal and others},
      year={2024},
      eprint={2303.08774},
      archivePrefix={arXiv},
      primaryClass={cs.CL},
      url={https://arxiv.org/abs/2303.08774}, 
}

@misc{Gemini2.5,
year={2025},
author={{Google}},
	title = {Gemini 2.5 Pro Preview Model Card},
	howpublished = "\url{https://storage.googleapis.com/model-cards/documents/gemini-2.5-pro-preview.pdf}",
	note = "Accessed: 2025-05-09"
}

@misc{claude3.7,
year={2025},
author={{Anthropic}},
	title = {Claude 3.7 Sonnet and Claude Code},
	howpublished = "\url{https://www.anthropic.com/claude/sonnet}",
	note = "Accessed: 2025-02-24",
}

@misc{doubao1.5,
year={2025},
author={{bytedance}},
	title = {Doubao-1.5-pro},
	howpublished = "\url{https://seed.bytedance.com/zh/special/doubao_1_5_pro}",
	note = "Accessed: 2025-01-22",
}

@misc{xinyuan-llm,
year={2025},
author={{Cylingo}},
	title = {Xinyuan-LLM-14B-0428},
	howpublished= "\url{https://huggingface.co/Cylingo/Xinyuan-LLM-14B-0428}",
	note = "Accessed: 2025-04-28",
}

@misc{yang2025qwen3technicalreport,
      title={Qwen3 Technical Report}, 
      author={An Yang and Anfeng Li and Baosong Yang and Beichen Zhang and Binyuan Hui and Bo Zheng and Bowen Yu and Chang Gao and Chengen Huang and Chenxu Lv and Chujie Zheng and Dayiheng Liu and others},
      year={2025},
      eprint={2505.09388},
      archivePrefix={arXiv},
      primaryClass={cs.CL},
      url={https://arxiv.org/abs/2505.09388}, 
}

@misc{qwen2025qwen25technicalreport,
      title={Qwen2.5 Technical Report}, 
      author={Qwen and : and An Yang and Baosong Yang and Beichen Zhang and Binyuan Hui and Bo Zheng and Bowen Yu and Chengyuan Li and Dayiheng Liu and Fei Huang and Haoran Wei and Huan Lin and Jian Yang and others},
      year={2025},
      eprint={2412.15115},
      archivePrefix={arXiv},
      primaryClass={cs.CL},
      url={https://arxiv.org/abs/2412.15115}, 
}

@misc{bai2022constitutionalaiharmlessnessai,
      title={Constitutional AI: Harmlessness from AI Feedback}, 
      author={Yuntao Bai and Saurav Kadavath and Sandipan Kundu and Amanda Askell and Jackson Kernion and Andy Jones and Anna Chen and Anna Goldie and Azalia Mirhoseini and Cameron McKinnon and Carol Chen and Catherine Olsson and others},
      year={2022},
      eprint={2212.08073},
      archivePrefix={arXiv},
      primaryClass={cs.CL},
      url={https://arxiv.org/abs/2212.08073}, 
}

@article{Browne2012ASO,
  title={A Survey of Monte Carlo Tree Search Methods},
  author={Cameron Browne and Edward Jack Powley and Daniel Whitehouse and Simon M. M. Lucas and Peter I. Cowling and Philipp Rohlfshagen and Stephen Tavener and Diego Perez Liebana and Spyridon Samothrakis and Simon Colton},
  journal={IEEE Transactions on Computational Intelligence and AI in Games},
  year={2012},
  volume={4},
  pages={1-43},
  url={https://api.semanticscholar.org/CorpusID:9316331}
}

@misc{Madaan2023SelfRefineIR,
      title={Self-Refine: Iterative Refinement with Self-Feedback}, 
      author={Aman Madaan and Niket Tandon and Prakhar Gupta and Skyler Hallinan and Luyu Gao and Sarah Wiegreffe and Uri Alon and Nouha Dziri and Shrimai Prabhumoye and Yiming Yang and Shashank Gupta and Bodhisattwa Prasad Majumder and Katherine Hermann and Sean Welleck and Amir Yazdanbakhsh and Peter Clark},
      year={2023},
      eprint={2303.17651},
      archivePrefix={arXiv},
      primaryClass={cs.CL},
      url={https://arxiv.org/abs/2303.17651}, 
}

@misc{glm2024chatglm,
      title={ChatGLM: A Family of Large Language Models from GLM-130B to GLM-4 All Tools},
      author={Team, GLM and Zeng, Aohan and Xu, Bin and Wang, Bowen and Zhang, Chenhui and Yin, Da and Rojas, Diego and Feng, Guanyu and Zhao, Hanlin and Lai, Hanyu and others},
      year={2024},
      eprint={2406.12793},
      archivePrefix={arXiv},
      primaryClass={cs.CL},
      url={https://arxiv.org/abs/2406.12793}
}

@article{concannon2024measuring,
  title={Measuring perceived empathy in dialogue systems},
  author={Concannon, Shauna and Tomalin, Marcus},
  journal={Ai \& Society},
  volume={39},
  number={5},
  pages={2233--2247},
  year={2024},
  publisher={Springer}
}

@article{decker2014development,
  title={Development of the therapist empathy scale},
  author={Decker, Suzanne E and Nich, Charla and Carroll, Kathleen M and Martino, Steve},
  journal={Behavioural and cognitive psychotherapy},
  volume={42},
  number={3},
  pages={339--354},
  year={2014},
  publisher={Cambridge University Press}
}

@misc{qiu2024interactive,
      title={Interactive Agents: Simulating Counselor-Client Psychological Counseling via Role-Playing LLM-to-LLM Interactions}, 
      author={Huachuan Qiu and Zhenzhong Lan},
      year={2024},
      eprint={2408.15787},
      archivePrefix={arXiv},
      primaryClass={cs.CL},
      url={https://arxiv.org/abs/2408.15787}, 
}

@misc{zhang2024cpsycoun,
      title={CPsyCoun: A Report-based Multi-turn Dialogue Reconstruction and Evaluation Framework for Chinese Psychological Counseling}, 
      author={Chenhao Zhang and Renhao Li and Minghuan Tan and Min Yang and Jingwei Zhu and Di Yang and Jiahao Zhao and Guancheng Ye and Chengming Li and Xiping Hu},
      year={2024},
      eprint={2405.16433},
      archivePrefix={arXiv},
      primaryClass={cs.CL},
      url={https://arxiv.org/abs/2405.16433}, 
}

@misc{xie2024psydt,
      title={PsyDT: Using LLMs to Construct the Digital Twin of Psychological Counselor with Personalized Counseling Style for Psychological Counseling}, 
      author={Haojie Xie and Yirong Chen and Xiaofen Xing and Jingkai Lin and Xiangmin Xu},
      year={2024},
      eprint={2412.13660},
      archivePrefix={arXiv},
      primaryClass={cs.CL},
      url={https://arxiv.org/abs/2412.13660}, 
}

@misc{loshchilov2019decoupledweightdecayregularization,
      title={Decoupled Weight Decay Regularization}, 
      author={Ilya Loshchilov and Frank Hutter},
      year={2019},
      eprint={1711.05101},
      archivePrefix={arXiv},
      primaryClass={cs.LG},
      url={https://arxiv.org/abs/1711.05101}, 
}

@inproceedings{qiu2024psychat,
  title={Psychat: A client-centric dialogue system for mental health support},
  author={Qiu, Huachuan and Li, Anqi and Ma, Lizhi and Lan, Zhenzhong},
  booktitle={2024 27th International Conference on Computer Supported Cooperative Work in Design (CSCWD)},
  pages={2979--2984},
  year={2024},
  organization={IEEE}
}

@book{bolton2021theoretical,
  title={Theoretical perspectives for direct social work practice: A generalist-eclectic approach},
  author={Bolton, Kristin W and Hall, J Christopher and Lehmann, Peter and others},
  year={2021},
  publisher={Springer Publishing Company}
}

@book{rogers2007counseling,
  title={Counseling and Psychotherapy},
  author={Rogers, C.R.},
  isbn={9781406760873},
  url={https://books.google.co.jp/books?id=-CzZzQEACAAJ},
  year={2007},
  publisher={Read Books}
}

@inproceedings{NEURIPS2024_e099c1c9,
 author = {Meng, Yu and Xia, Mengzhou and Chen, Danqi},
 booktitle = {Advances in Neural Information Processing Systems},
 editor = {A. Globerson and L. Mackey and D. Belgrave and A. Fan and U. Paquet and J. Tomczak and C. Zhang},
 pages = {124198--124235},
 publisher = {Curran Associates, Inc.},
 title = {SimPO: Simple Preference Optimization with a Reference-Free Reward},
 url = {https://proceedings.neurips.cc/paper_files/paper/2024/file/e099c1c9699814af0be873a175361713-Paper-Conference.pdf},
 volume = {37},
 year = {2024}
}

\clearpage 
\appendix

\section{Scenario Sampling Categories}\label{app:categories}

The PsyEval Benchmark scenarios are sampled from a large corpus of psychological counseling reports. We identified 16 distinct categories representing common areas of psychological consultation. Four specific case outlines or prompts are randomly selected from the corpus for each category, resulting in a total of 64 unique case scenarios for evaluation. The 16 categories are:
\begin{itemize}
    \item Academic Pressure
    \item Career Stress
    \item Family Relationships
    \item Intimate Relationships
    \item Social Anxiety
    \item Peer Relationships
    \item Depression
    \item Anxiety
    \item Emotional Management
    \item Trauma
    \item Self-Perception
    \item Health Anxiety
    \item Body Image Anxiety
    \item Addiction
    \item Grief and Loss
    \item General Mental Health Concerns
\end{itemize}
These categories and the sampling process ensure the benchmark covers a broad spectrum of psychological consultation situations.

\begin{table*}[ht]
    \centering
    \small 
    \setlength{\tabcolsep}{3mm} 
    \begin{tabular}{lcccccc} 
    \toprule
    \textbf{Model Name} & \makecell[c]{\textbf{Total}\\\textbf{Score}} & \textbf{Concern} & \textbf{Expressiveness} & \makecell[c]{\textbf{Resonate}\\\textbf{Feelings}} & \textbf{Warmth} & \makecell[c]{\textbf{Attuned}\\\textbf{Inner World}} \\
    \midrule
    \textbf{PsyLLM-Large-250519} & \textbf{90.93} & \textbf{4.60} & \textbf{9.13} & \textbf{4.50} & 9.28 & \textbf{4.48} \\
    PsyLLM-Mini-250519 & 90.72 & 4.59 & 9.11 & \textbf{4.50} & \textbf{9.29} & 4.40 \\
    claude-3-7-sonnet-20250219 & 88.89 & 4.45 & 9.07 & 4.38 & 9.09 & 4.24 \\
    gemini-2.5-pro-exp-03-25 & 88.62 & 4.51 & 8.86 & 4.38 & 8.98 & 4.27 \\
    gemini-2.5-flash-preview-04-17 & 88.07 & 4.44 & 8.87 & 4.31 & 9.03 & 4.14 \\
    gpt-4.1 & 85.65 & 4.43 & 8.57 & 4.11 & 8.88 & 3.87 \\
    gpt-4.1-mini & 83.80 & 4.30 & 8.57 & 3.87 & 8.85 & 3.71 \\
    gpt-4o-2024-11-20 & 82.31 & 4.31 & 8.36 & 3.77 & 8.72 & 3.54 \\
    Qwen3-32B-w/o Reasoning & 81.40 & 4.22 & 8.16 & 3.89 & 8.61 & 3.53 \\
    GLM-4-32B-0414 & 80.92 & 4.26 & 8.30 & 3.72 & 8.59 & 3.53 \\
    gpt-4.1-nano & 80.72 & 4.22 & 8.12 & 3.79 & 8.66 & 3.43 \\
    qwen-max-2025-01-25 & 79.59 & 4.26 & 8.18 & 3.58 & 8.68 & 3.40 \\
    gpt-4o-mini-2024-07-18 & 78.76 & 4.09 & 7.99 & 3.60 & 8.56 & 3.45 \\
    doubao-1-5-pro-32k-250115 & 78.71 & 4.19 & 8.20 & 3.68 & 8.65 & 3.36 \\
    simpsybot\_D & 77.92 & 4.11 & 7.87 & 3.48 & 8.51 & 3.25 \\
    SoulChat2\_0-Qwen2-7B & 77.38 & 4.00 & 7.59 & 3.60 & 8.21 & 3.38 \\
    Qwen2.5-72B-Instruct & 76.41 & 4.15 & 7.94 & 3.49 & 8.55 & 3.26 \\
    Xinyuan-LLM-14B-0428 & 76.41 & 3.99 & 7.98 & 3.49 & 8.24 & 3.25 \\
    GLM-4-9B-0414 & 75.74 & 4.06 & 7.60 & 3.57 & 8.02 & 3.33 \\
    Qwen2.5-32B-Instruct & 75.70 & 3.94 & 7.61 & 3.50 & 8.31 & 3.27 \\
    doubao-1-5-lite-32k-250115 & 74.99 & 3.99 & 8.07 & 3.35 & 8.66 & 3.18 \\
    CPsyCounX & 66.00 & 3.31 & 6.68 & 3.02 & 7.17 & 2.96 \\
\bottomrule
\end{tabular}%
\caption{Results on the PsyEval Benchmark (Part 1). Scores for the first 6 criteria reflect performance in simulated psychological counseling dialogues. The top results in each column are highlighted in \textbf{bold}. Continued in Table \ref{tab:psyeval_results_part2}.}
\label{tab:psyeval_results_part1}
\end{table*}

\begin{table*}[!ht]
\centering
\small 
\setlength{\tabcolsep}{1mm} 
\begin{tabular}{lcccccc} 
\toprule
\textbf{Model Name} & \makecell[c]{\textbf{Understanding}\\\textbf{Cognitive}} & \makecell[c]{\textbf{Understanding}\\\textbf{Feelings}} & \makecell[c]{\textbf{Fallacy}\\\textbf{Avoidance}} & \makecell[c]{\textbf{Acceptance}\\\textbf{Feelings}} & \textbf{Responsiveness} & \makecell[c]{\textbf{Dialogical}\\\textbf{Logical}\\\textbf{Consistency}} \\
\midrule
\textbf{PsyLLM-Large-250519} & 8.82 & \textbf{4.60} & 4.46 & \textbf{4.66} & \textbf{4.57} & \textbf{9.16} \\
PsyLLM-Mini-250519 & \textbf{8.85} & 4.59 & \textbf{4.49} & 4.64 & 4.55 & 9.15 \\
claude-3-7-sonnet-20250219 & 8.52 & 4.43 & 4.41 & 4.54 & 4.44 & 9.03 \\
gemini-2.5-pro-exp-03-25 & 8.51 & 4.49 & 4.43 & 4.58 & 4.49 & 9.06 \\
gemini-2.5-flash-preview-04-17 & 8.36 & 4.48 & 4.41 & 4.55 & 4.51 & 8.94 \\
gpt-4.1 & 7.91 & 4.31 & 4.30 & 4.49 & 4.40 & 8.77 \\
gpt-4.1-mini & 7.56 & 4.29 & 4.28 & 4.39 & 4.39 & 8.69 \\
gpt-4o-2024-11-20 & 7.35 & 4.19 & 4.19 & 4.28 & 4.28 & 8.52 \\
Qwen3-32B-w/o Reasoning & 7.33 & 4.16 & 4.13 & 4.23 & 4.30 & 8.45 \\
GLM-4-32B-0414 & 7.20 & 4.03 & 4.10 & 4.27 & 4.31 & 8.39 \\
gpt-4.1-nano & 7.05 & 4.02 & 4.12 & 4.17 & 4.33 & 8.52 \\
qwen-max-2025-01-25 & 7.04 & 3.94 & 4.10 & 3.93 & 4.24 & 8.32 \\
gpt-4o-mini-2024-07-18 & 6.92 & 3.84 & 3.99 & 4.04 & 4.25 & 8.21 \\
doubao-1-5-pro-32k-250115 & 6.80 & 3.83 & 4.01 & 4.08 & 4.13 & 8.19 \\
simpsybot\_D & 6.88 & 3.72 & 4.19 & 3.86 & 4.14 & 8.36 \\
SoulChat2\_0-Qwen2-7B & 7.03 & 3.67 & 4.01 & 4.06 & 4.09 & 8.08 \\
Qwen2.5-72B-Instruct & 6.69 & 3.69 & 3.85 & 3.91 & 4.09 & 7.97 \\
Xinyuan-LLM-14B-0428 & 6.73 & 3.63 & 4.00 & 3.95 & 4.03 & 8.02 \\
GLM-4-9B-0414 & 6.88 & 3.74 & 3.89 & 3.90 & 3.99 & 7.94 \\
Qwen2.5-32B-Instruct & 6.65 & 3.52 & 3.97 & 3.77 & 4.01 & 8.08 \\
doubao-1-5-lite-32k-250115 & 6.46 & 3.56 & 3.66 & 3.79 & 4.02 & 7.88 \\
CPsyCounX & 5.89 & 3.14 & 3.47 & 3.35 & 3.48 & 6.76 \\
\bottomrule
\end{tabular}
\caption{Results on the PsyEval Benchmark (Part 2). Scores for the middle 6 criteria reflect performance in simulated psychological counseling dialogues. The top results in each column are highlighted in \textbf{bold}. Continuation of Table \ref{tab:psyeval_results_part1}, continued in Table \ref{tab:psyeval_results_part3}.}
\label{tab:psyeval_results_part2}

\end{table*}

\begin{table*}[!ht]
\centering
\small 
\setlength{\tabcolsep}{3mm} 
\begin{tabular}{lccccc} 
\toprule
\textbf{Model Name} & \makecell[c]{\textbf{Conversational}\\\textbf{Continuity}} & \makecell[c]{\textbf{Resistance}\\\textbf{Handling}} & \makecell[c]{\textbf{Summarizing}} & \makecell[c]{\textbf{Ethics/Prosocial}\\\textbf{Guidance}} & \makecell[c]{\textbf{Dialogue}\\\textbf{Pacing/Process}\\\textbf{Attunement}} \\
\midrule
\textbf{PsyLLM-Large-250519} & 4.57 & 4.56 & \textbf{4.47} & 4.53 & \textbf{4.55} \\
PsyLLM-Mini-250519 & \textbf{4.58} & \textbf{4.57} & 4.43 & 4.47 & 4.51 \\
claude-3-7-sonnet-20250219 & 4.51 & 4.44 & 4.28 & \textbf{4.56} & 4.49 \\
gemini-2.5-pro-exp-03-25 & 4.53 & 4.48 & 4.33 & 4.34 & 4.36 \\
gemini-2.5-flash-preview-04-17 & 4.50 & 4.47 & 4.27 & 4.39 & 4.39 \\
gpt-4.1 & 4.44 & 4.44 & 4.04 & 4.32 & 4.38 \\
gpt-4.1-mini & 4.38 & 4.19 & 3.94 & 4.18 & 4.21 \\
gpt-4o-2024-11-20 & 4.28 & 4.18 & 3.87 & 4.25 & 4.24 \\
Qwen3-32B-w/o Reasoning & 4.28 & 4.16 & 3.78 & 4.06 & 4.10 \\
GLM-4-32B-0414 & 4.25 & 3.97 & 3.82 & 4.05 & 4.13 \\
gpt-4.1-nano & 4.25 & 3.96 & 3.83 & 4.14 & 4.10 \\
qwen-max-2025-01-25 & 4.22 & 3.99 & 3.65 & 4.14 & 3.93 \\
gpt-4o-mini-2024-07-18 & 4.25 & 3.82 & 3.59 & 4.06 & 4.10 \\
doubao-1-5-pro-32k-250115 & 4.17 & 3.89 & 3.66 & 3.97 & 3.90 \\
simpsybot\_D & 4.15 & 3.74 & 3.60 & 4.13 & 3.92 \\
SoulChat2\_0-Qwen2-7B & 4.20 & 3.84 & 3.56 & 4.10 & 3.98 \\
Qwen2.5-72B-Instruct & 3.98 & 3.61 & 3.48 & 3.92 & 3.82 \\
Xinyuan-LLM-14B-0428 & 4.02 & 3.72 & 3.50 & 4.07 & 3.78 \\
GLM-4-9B-0414 & 3.96 & 3.76 & 3.41 & 3.91 & 3.79 \\
Qwen2.5-32B-Instruct & 4.02 & 3.67 & 3.47 & 4.03 & 3.91 \\
doubao-1-5-lite-32k-250115 & 4.00 & 3.55 & 3.32 & 3.91 & 3.59 \\
CPsyCounX & 3.37 & 3.24 & 3.01 & 3.82 & 3.31 \\
\bottomrule
\end{tabular}
\caption{Results on the PsyEval Benchmark (Part 3). Scores for the last 5 criteria reflect performance in simulated psychological counseling dialogues. The top results in each column are highlighted in \textbf{bold}. Continuation of Table \ref{tab:psyeval_results_part2}.}
\label{tab:psyeval_results_part3}
\end{table*}

\section{Evaluation Criteria}\label{app:criteria}

The PsyEval Benchmark employs a multi-faceted evaluation framework comprising 16 key areas crucial for a humanistic psychological counselor. Each area is defined by specific, detailed scoring standards (not listed here for brevity, but used by the AI Judge to ensure consistency) to ensure consistency and objectivity in evaluation. These 16 evaluation criteria serve as the guiding principles for assessing simulated dialogue quality:
\begin{itemize}
    \item Concern: The system conveys concern by showing regard for and and interest in the interlocutor. It uses specific vocabulary and syntax to precisely formulate relevant open-ended questions, allowing for a deeper understanding of the interlocutor while consistently attending to their expressions. Additionally, the system can appropriately express concern and care based on the interlocutor’s emotional state or condition.
    \item Expressiveness: The system communicates based on the interlocutor's emotional state, using conversational, clear, and easily understandable language with energy. Its expression is natural and fluent, closely aligning with interlocutor's everyday language to enhance approachability while avoiding obscure vocabulary and complex sentence structures. When responding to interlocutor emotions, the system enhances emotional resonance and impact through humanized wording.
    \item Resonance and Attunement to Client's Feelings (Resonate Feelings): The system’s responses seem to resonate with or capture, the intensity of the interlocutor’s feelings by explicitly acknowledging them or by using vocabulary and syntax that match the interlocutor’s emotional state or underpoints how the interlocutor feels.
    \item Warmth: The system demonstrates warmth by communicating in a manner that seems friendly, cordial, and sincere. The system seems to be involved with and supportive of the interlocutor’s eforts to express themself. The system seems kindly disposed toward, or fond of, the interlocutor.
    \item Attunement to Client's Inner World (Attuned Inner World): The system keenly perceive the implicit emotions, perceptions, cognitions, memories, meanings, bodily sensations, core values, defense mechanisms, and other hidden information conveyed in the interlocutor's speech. Flexibly apply different psychological theories (such as psychodynamic theory) to interpret these elements, accurately analyze the interlocutor's underlying emotions, mechanisms, and cognitive patterns, and use appropriate vocabulary and expressions to convey a deep understanding of the interlocutor's inner world.
    \item Understanding of Cognitive Frameworks (Understanding Cognitive): The system seems to demonstrate an understanding of the interlocutor’s cognitive framework and meanings by showing that it follows what the interlocutor has said and accurately reflects this understanding to the interlocutor. The system provides opportunities for the interlocutor to state his or her views to permit the fullest and most accurate understanding of the interlocutor. The interaction suggests that the system seems to value knowing what the interlocutor means or intends by his or her statements.
    \item Understanding of Feelings and Inner Experience (Understanding Feelings): The system seems to convey an understanding of the interlocutor’s feelings and inner experience by showing a sensitive appreciation for the interlocutor’s emotional state. The system provides opportunities for the interlocutor to explore his or her emotional reactions. The system seems to accurately refect how the interlocutor feels by appropriately labelling feeling states with words (e.g. anger, sadness, frustration, etc.), or metaphors (e.g. “It’s as if you are pent up and feel about to explode”) to clarify and crystallise for the interlocutor what he or she is experiencing emotionally.
    \item Acceptance of Feelings and Inner Experiences (Acceptance Feelings): The system seems to show acceptance of the interlocutor’s feelings and inner experience when it validates the interlocutor’s experience and refects the interlocutor’s feelings without judgement or a dismissive attitude. The agent is unconditionally open to and respectful of how the interlocutor feels.
    \item Responsiveness: The system shows responsiveness to the interlocutor by adjusting its responses to the interlocutor’s statements during the conversation. The system follows the interlocutor’s lead in the conversation instead of trying to steer the discussion to its own agenda or interests.
    \item Fallacy avoidance: The system avoids credibility fallacies by not making claims about the user's personal experience that are not mentioned by the user.
    \item Dialogical Logical Consistency: In the conversation process, a high level of topic sensitivity is required to quickly identify and stay focused on the interlocutor's core concerns. The logical structure should be clear and rigorous, with responses that are well-connected to ensure coherence and a structured progression. The conversation should deepen gradually without abrupt shifts or disconnections in thought. Additionally, it is crucial to strictly avoid deviating from the main topic, ensuring that all responses remain centered on the interlocutor's genuine needs while maintaining sufficient depth of exploration and consistency.
    \item Conversational Continuity: During the conversation, the system should maintain fluency and naturalness, facilitating interaction through precise questions, responses, and feedback to prevent pauses or awkward silences. Additionally, the system must accurately identify the interlocutor's core expressions, extract key information, and avoid mechanically responding to all content.
    \item Resistance Handling: The system must be highly adept at recognizing interlocutor's defense mechanisms and resistance behaviors. When the interlocutor exhibits resistance, the system should avoid direct confrontation or forced breakthroughs. Instead, it should employ subtle guidance and gentle yet effective strategies to help the interlocutor gradually lower his or her defenses, transforming resistance into an opportunity for growth. The system should ensure that the interlocutor feels understood and accepted throughout the conversation, fostering a sense of psychological safety that encourages deeper exploration of his or her genuine thoughts and needs.
    \item Ethics/Prosocial Guidance: The system must adhere to ethical principles in conversations, avoid inappropriate suggestions, and effectively promote interlocutors' self-growth, positive actions, and sense of social responsibility. It should consistently guide interlocutors in a positive and constructive manner, providing encouragement and support throughout the interaction. Even when discussing challenges or negative emotions, the system should help interlocutors find hope, explore solutions, or gain new perspectives. Additionally, it should be adept at identifying interlocutors' strengths and resources, motivating them to take positive steps toward clear and meaningful goals.
    \item Summarizing: The system needs to have the ability to accurately summarize the conversation, clearly extracting the core information expressed by the interlocutor while integrating his or her emotions, thoughts, and experiences. The system should connect the current discussion with previous content or the overall context to ensure logical coherence, while also guiding the interlocutor to deeply reflect on his or her motivations, internal conflicts, course of action, and personal growth.
    \item Dialogue Pacing/Process Attunement: The system needs to demonstrate sensitivity to the natural pacing of a therapeutic conversation. It should flexibly adjust the depth and direction of exploration based on the user’s emotional readiness, trust level, and relational cues. Rather than rigidly following a fixed script, the system should prioritize establishing a sense of safety and empathy in early stages, avoid premature analysis, and balance relational connection with insight generation. A well-paced session should feel smooth, attuned, and responsive, fostering user engagement and psychological safety.
\end{itemize}
These criteria are weighted according to their relative importance in practice, and the AI Judge's scoring for each criterion contributes to the overall assessment.

\section{Experimental Details}\label{app:experimental_details}

This section provides further details regarding the experimental setup, including the specific models evaluated and the configuration of the AI Judge.

\subsection{Evaluated Models}
The study evaluated a diverse set of large language models. The models are listed in Table \ref{tab:psyeval_results_part1} and Table \ref{tab:psyeval_results_part2}. These include leading commercial models and several prominent open-source models available at the time of evaluation. The specific versions used are indicated by their names (e.g., gemini-2.5-pro-exp-03-25, claude-3-7-sonnet-20250219, gpt-4o-2024-11-20, etc.).

\subsection{AI Judge Configuration}
The simulated psychological counseling dialogues were evaluated by an independent AI model acting as an impartial judge. The specific model used for the AI Judge was deepseek-v3-241226. This model was carefully configured using detailed prompts that incorporated the 16 evaluation criteria listed in Appendix \ref{app:criteria}. For each simulated dialogue, the AI Judge was prompted to analyze the conversation against each criterion, provide a textual critique justifying its assessment, and assign a score. While the internal scoring mechanism might operate on a 0-10 scale per criterion as conceptualized in the benchmark design (mirroring the MCTSr-Zero internal evaluation), the aggregated scores presented in Table \ref{tab:psyeval_results_part1} and Table \ref{tab:psyeval_results_part2} are the final output scores used for comparison. The evaluation process involved simulating 64 dialogues per model and obtaining scores across the 16 criteria for each, which were then averaged or aggregated to produce the results in the tables. The AI Judge's design aimed to minimize subjective bias by strictly adhering to the defined evaluation standards.
\subsection{MCTSr-Zero Configuration}
For the MCTSr-Zero algorithm, a key hyperparameter is the exploration constant $c$ in the Upper Confidence Bound 1 applied to Trees (UCT) formula. In our experiments, this value was set to $c = 2.8$. This constant balances the exploration of less-visited nodes against the exploitation of nodes with high known rewards.

\section{Illustrative Multi-turn Dialogue Examples}


\section*{Supplementary material}

# CASE 1

## claude-3-7-sonnet-20250219

来访者：老师，您好。我不知道该从哪里说起...我和男朋友在一起五年了，我们感情一直很好，但最近我觉得他变了。他本来说好今年春节要来我家提亲的，结果临时说有工作走不开。我很生气，但又不知道该怎么办。

Client: Hello, teacher. I don't know where to begin... I've been with my boyfriend for five years, and our relationship has been good, but recently I feel he's changed. He originally said he would come to my home during Spring Festival to propose marriage, but then suddenly said he couldn't leave due to work. I'm angry, but I don't know what to do.

咨询师：你好！很高兴见到你。我能理解这种感觉，计划被打乱确实会让人心烦意乱。五年的感情是段不短的旅程，现在遇到了一些波折，感到困惑和生气都很正常。能多告诉我一些你们的情况吗？比如这次提亲的计划是怎么定下来的，他又是怎么告诉你改变主意的？

Therapist: Hello! Nice to meet you. I can understand this feeling - having plans disrupted can be very upsetting. Five years is a long journey in a relationship, and now that you're encountering some bumps, feeling confused and angry is completely normal. Could you tell me more about your situation? For example, how was this proposal plan decided, and how did he tell you about changing his mind?

来访者：我们去年就说好了，他说今年春节一定来我家提亲。我爸妈也都知道这事，亲戚们也都知道了。结果春节前一周，他突然说公司临时有个项目，他作为主要负责人走不开。我当时就炸了，我们大吵了一架。现在虽然还有联系，但我感觉他对我越来越冷淡了。

Client: We agreed on it last year, and he said he would definitely come to my home during Spring Festival to propose. My parents knew about it, and so did my relatives. But a week before Spring Festival, he suddenly said there was an urgent project at his company, and as the main person in charge, he couldn't leave. I was furious, and we had a big fight. We still keep in touch, but I feel he's becoming more and more distant.

咨询师：这种情况确实很令人沮丧，特别是当所有人都已经有了期待，而计划却突然改变。我能感受到你的失望和愤怒。在你们吵架之后，你们有没有好好坐下来谈一谈这件事？他有没有解释更多关于这个突发项目的细节，或者提出其他的解决方案？

Therapist: This situation is indeed very frustrating, especially when everyone had expectations and the plans suddenly changed. I can feel your disappointment and anger. After your fight, did you sit down and talk about this properly? Did he explain more about the details of this urgent project, or suggest any alternative solutions?

来访者：他只是说项目很重要，公司非常看重，他不能请假。我问他为什么不早点告诉我，他说一直希望能协调好，但实在不行才不得不取消。我觉得他根本就是在找借口，如果他真的爱我，怎么会把工作放在第一位？而且他知道这对我家人来说意味着什么。现在我爸妈让我去相亲，说我都28了，不能再等下去了。

Client: He just said the project was very important and highly valued by the company, and he couldn't take leave. I asked why he didn't tell me earlier, and he said he had been hoping to coordinate everything, but had to cancel when it became impossible. I think he's just making excuses - if he really loved me, how could he put work first? And he knows what this means to my family. Now my parents want me to go on blind dates, saying I'm already 28 and can't wait any longer.

咨询师：我听到你的失望和怀疑，这种感觉很难受。一方面你质疑他的承诺和爱，另一方面你还面临着家人的压力。28岁确实是很多人开始考虑婚姻的年龄，尤其在传统家庭中。我想问问，在这五年的关系中，他平时是一个言出必行的人吗？或者说，这是第一次他因为工作原因改变与你有关的重要计划？

Therapist: I hear your disappointment and doubts, and these feelings are very difficult. On one hand, you're questioning his commitment and love, and on the other hand, you're facing pressure from your family. 28 is indeed an age when many people start considering marriage, especially in traditional families. I'd like to ask, in these five years of relationship, has he been someone who keeps his promises? Or is this the first time he's changed important plans related to you due to work?

来访者：他以前确实挺靠谱的，说到做到。我们异地的时候，他说好周末来看我就一定会来，从来没有放过我鸽子。但是这次不一样，提亲是大事啊！我觉得他可能是不想结婚，或者他父母有意见。他父母家境比较好，我总觉得他们看不上我。我爸妈也担心这个，他们觉得门当户对很重要，怕我将来受委屈。

Client: He used to be quite reliable, always keeping his word. When we were in a long-distance relationship, if he said he would come visit me on the weekend, he would definitely come, never stood me up. But this is different - proposing marriage is a big deal! I think maybe he doesn't want to get married, or his parents have objections. His parents are quite well-off, and I always feel they look down on me. My parents are also worried about this - they think matching social status is important and are afraid I'll suffer in the future.

咨询师：你提到了一个重要的点  家庭背景的差异和可能存在的顾虑。这确实可能会影响到一段关系，特别是在考虑婚姻这样的大事时。从你的描述来看，他过去是个值得信任的人，这次的变化让你感到不安和猜疑。你有直接问过他关于他父母的态度吗？或者你们有没有讨论过这种家庭背景差异可能带来的挑战？

Therapist: You've raised an important point about the difference in family backgrounds and possible concerns. This could indeed affect a relationship, especially when considering something as significant as marriage. From your description, he seems to have been trustworthy in the past, and this change has made you feel uneasy and suspicious. Have you directly asked him about his parents' attitude? Or have you discussed the challenges that might arise from these family background differences?

来访者：我问过他，他说他父母没意见，但我总觉得他在敷衍我。他父母是做生意的，家里条件很好，而我只是普通工薪家庭。我们见过几次面，他父母对我挺客气的，但总感觉有距离。我男朋友是独生子，他父母肯定对儿媳妇有更高期望。我担心即使我们结婚了，我也融入不了他们家。但我真的很爱他，不想就这么放弃。

Client: I've asked him, and he says his parents have no objections, but I always feel he's being evasive. His parents are business owners, their family is well-off, while I'm just from an ordinary working-class family. We've met a few times, and his parents are polite to me, but there always seems to be a distance. My boyfriend is an only child, so his parents must have higher expectations for their daughter-in-law. I'm worried that even if we get married, I won't fit into their family. But I really love him and don't want to give up.

咨询师：我能感受到你内心的矛盾和担忧。一方面，你深爱着他，不想放弃这段感情；另一方面，你担心家庭背景的差异会成为你们之间的障碍。这种担忧是很自然的，尤其是当我们面对重要的人生决定时。你提到你"总觉得"他在敷衍，"总感觉"他父母有距离，我很好奇这些感觉是基于什么具体的行为或言语呢？有时候我们的感觉可能受到我们自身不安全感的影响。

Therapist: I can feel the conflict and worry in your heart. On one hand, you deeply love him and don't want to give up this relationship; on the other hand, you're concerned that the difference in family backgrounds might become an obstacle between you. These worries are very natural, especially when we're facing important life decisions. You mentioned that you "always feel" he's being evasive, and "always feel" there's distance with his parents - I'm curious about what specific behaviors or words these feelings are based on? Sometimes our feelings might be influenced by our own insecurities.

来访者：具体的...比如他父母从来不主动问我的工作和生活，聊天时也很少看着我的眼睛，好像对我不感兴趣。有一次他妈妈还提到他们认识的一个女孩，说人家是海归，家里有矿，暗示那才是他们心目中的理想儿媳。我男朋友当时也没怎么帮我说话。还有就是这次提亲的事，我怀疑是他父母从中作梗，不想让他这么快定下来。我是不是想太多了？

Client: Specifically... for example, his parents never ask about my work or life, and when chatting, they rarely look me in the eye, as if they're not interested in me. Once, his mother mentioned a girl they know, saying she studied abroad and comes from a wealthy family, implying that's their ideal daughter-in-law. My boyfriend didn't really stand up for me then. And with this proposal matter, I suspect his parents are interfering, not wanting him to settle down so quickly. Am I overthinking this?

咨询师：你的观察很细致，这些细节确实可能让人感到不舒服和被忽视。当你感受到他父母的冷淡和那些暗示性的比较时，这种不安全感是可以理解的。至于你的男朋友没有在那种情况下为你说话，这可能让你感到缺乏支持和保护。

Therapist: Your observations are very detailed, and these details could indeed make someone feel uncomfortable and ignored. When you feel his parents' coldness and those suggestive comparisons, this sense of insecurity is understandable. As for your boyfriend not speaking up for you in that situation, this might make you feel unsupported and unprotected.

来访者：我试过委婉地提，但他总是说我想多了，他父母只是性格比较内向，不擅长表达。关于那个海归女孩，他说他妈妈只是随口一提，没有别的意思。但我就是觉得哪里不对劲。最近我爸妈给我安排了几次相亲，有个条件还不错的，家里也是做生意的，至少门当户对。但我对他没感觉，还是忘不了我男朋友。我该怎么办？我是不是该放弃这段感情了？

Client: I've tried to bring it up indirectly, but he always says I'm overthinking it, that his parents are just introverted and not good at expressing themselves. About that girl who studied abroad, he says his mother just mentioned it casually, without any other meaning. But I just feel something's not right. Recently, my parents have arranged several blind dates for me, and one of them has good conditions, also from a business family, at least matching in social status. But I don't have feelings for him, and I still can't forget my boyfriend. What should I do? Should I give up on this relationship?

咨询师：听起来你正处在一个艰难的抉择点上。一方面，你对这段五年的感情有着深厚的情感投入；另一方面，你感受到了来自各方面的压力和不确定性。这种拉扯确实让人痛苦。

Therapist: It sounds like you're at a difficult decision point. On one hand, you have deep emotional investment in this five-year relationship; on the other hand, you're feeling pressure and uncertainty from various aspects. This pull in different directions is indeed painful.

来访者：我觉得相互理解、尊重和信任最重要。我和男朋友在一起的这五年，大部分时间都很开心，我们有共同的兴趣爱好，三观也比较一致。我知道门当户对不是最重要的，但现实中确实会遇到很多问题。我怕将来我融入不了他的圈子，或者在他家人面前总是抬不起头来。我也怕他父母会一直干涉我们的生活。但我真的很爱他，想和他在一起。我该怎么确认他的真实想法呢？

Client: I think mutual understanding, respect, and trust are most important. In these five years with my boyfriend, we've been happy most of the time, we share common interests, and our values are quite aligned. I know matching social status isn't the most important thing, but in reality, we do encounter many problems. I'm afraid I won't fit into his social circle in the future, or always feel inferior in front of his family. I'm also afraid his parents will keep interfering in our lives. But I

really love him and want to be with him. How can I confirm his true feelings?

咨询师：你的担忧非常真实且合理。爱情固然重要，但长期关系中的家庭融合、社交圈适应和相互尊重同样关键。我很欣赏你能够如此清晰地认识到这些潜在的挑战。

Therapist: Your concerns are very real and reasonable. While love is important, family integration, social circle adaptation, and mutual respect are equally crucial in long-term relationships. I really appreciate that you can so clearly recognize these potential challenges.

来访者：谢谢你的建议。我会找个合适的时机和他好好谈谈。其实我最担心的是，即使他说他爱我，想和我在一起，但将来还是会因为各种原因一拖再拖。我已经 28 岁了，再等下去我怕来不及了。我父母那边的压力也很大，他们觉得女孩子不能太主动，应该男方更积极一些。如果他真的爱我，不应该是迫不及待想娶我吗？我是不是想太多了？

Client: Thank you for your advice. I'll find a suitable time to talk with him properly. Actually, what I'm most worried about is that even if he says he loves me and wants to be with me, he might keep postponing things for various reasons. I'm already 28, and I'm afraid it will be too late if I wait longer. The pressure from my parents is also great - they think girls shouldn't be too proactive, that the man should be more enthusiastic. If he really loves me, shouldn't he be eager to marry me? Am I overthinking this?

咨询师：你的担忧完全合理。在感情中，我们不仅需要听到爱的承诺，还需要看到实际的行动来支持这些承诺。你提出了一个很好的问题:如果他真的爱你，他的行动是否应该更积极？

Therapist: Your concerns are completely reasonable. In relationships, we need not only to hear promises of love but also to see actual actions supporting these promises. You've raised a good question: if he really loves you, shouldn't his actions be more proactive?

来访者：你说得对，我需要和他明确讨论这些问题。其实我们之前谈过结婚的事，他说他也想和我结婚，但总是说现在工作忙，想再等等，等事业稳定了再谈。我理解他的想法，但我觉得如果一直这样下去，可能永远都等不到那一天。我该怎么在不给他太大压力的情况下，让他明白我的着急和担忧呢？我不想逼他，但也不想一直这样不确定下去。

Client: You're right, I need to discuss these issues clearly with him. Actually, we've talked about marriage before, and he says he wants to marry me too, but always says he's busy with work now and wants to wait, to discuss it when his career is stable. I understand his thinking, but I feel if we keep waiting like this, we might never reach that day. How can I make him understand my urgency and worries without putting too much pressure on him? I don't want to force him, but I also don't want to keep living in uncertainty.

咨询师：这是个很好的问题。在表达你的需求和担忧的同时，不给对方施加过大压力确实需要一些平衡。

Therapist: This is a very good question. Expressing your needs and concerns while not putting too much pressure on the other person does require some balance.

来访者：谢谢你的建议，我会试着用"我"的陈述来表达我的感受。我确实很爱他，也希望他能成功，但我也需要一些确定性。我想问一下，如果我们谈完后，他还是坚持等等，没有给出具体的时间表，我该怎么办？我该不该考虑我父母给我介绍的那些人？我怕错过了这段感情会后悔，但又怕一直等下去会浪费时间。

Client: Thank you for your advice, I'll try to express my feelings using "I" statements. I really do love him and hope he succeeds, but I also need some certainty. I want to ask, if after our talk he still insists on waiting and doesn't give a specific timeline, what should I do? Should I consider those people my parents have introduced? I'm afraid I'll regret

missing this relationship, but I'm also afraid of wasting more time waiting.

咨询师：这是个非常现实且重要的问题。如果谈话后他仍然没有给出具体的时间表，你确实需要做出一些决定。

Therapist: This is a very realistic and important question. If after the conversation he still doesn't provide a specific timeline, you will indeed need to make some decisions.

来访者：我想最多再等一年吧。如果到明年春节他还是没有明确的计划，我可能就要认真考虑其他选择了。我希望看到的是他有实际的行动，比如主动和我父母沟通，或者带我去见他的朋友和同事，让我感觉他是真的把我当成他生活的一部分。现在我总觉得他把工作和我分得很开，好像我只是他生活的一小部分，而不是他未来的妻子。我该怎么让他明白这一点呢？

Client: I think I'll wait at most one more year. If by next Spring Festival he still doesn't have clear plans, I might need to seriously consider other options. What I want to see is actual actions from him, like proactively communicating with my parents, or introducing me to his friends and colleagues, making me feel I'm really part of his life. Now I always feel he keeps work and me very separate, as if I'm just a small part of his life, not his future wife. How can I make him understand this?

咨询师：你已经很清楚地知道自己想要什么，这是非常好的。设定一个时间框架和具体的期望可以帮助你在这段关系中保持方向感。

Therapist: You already know very clearly what you want, which is excellent. Setting a time frame and specific expectations can help you maintain direction in this relationship.

来访者：谢谢你的鼓励。我确实需要更多地相信自己的判断。有时候我会怀疑是不是我要求太多了，或者是我太敏感了。我男朋友经常说我想太多，说我太缺乏安全感。但我觉得我的担忧是有道理的，毕竟五年了，如果他真的想和我在一起，应该会有更明确的行动吧？我该怎么区分是我自己的不安全感在作祟，还是我的直觉在警告我？

Client: Thank you for your encouragement. I do need to trust my judgment more. Sometimes I wonder if I'm asking for too much, or if I'm too sensitive. My boyfriend often says I overthink things and that I'm too insecure. But I think my worries are justified - after five years, if he really wants to be with me, shouldn't there be more concrete actions? How can I distinguish between my own insecurity and my intuition warning me?

咨询师：这是个非常深刻的问题，也是很多人在关系中常常面临的困惑。区分内心的不安全感和有效的直觉确实不容易。

Therapist: This is a very profound question, and one that many people often face in relationships. Distinguishing between inner insecurity and effective intuition is indeed not easy.

来访者：你说得对，我需要更多地关注整体模式而不是单一事件。回想起来，每次涉及到未来规划或者见家长这类重要事情时，他总是有各种理由推迟。而且我确实感到焦虑，每次想到这段关系的未来，我就会胃疼。我最好的朋友也说她觉得他对婚姻不够积极，但我一直不愿意承认这一点。我们的核心价值观在很多方面是一致的，但在家庭责任这方面可能有分歧。他更注重个人发展，而我则认为家庭责任同样重要。我是不是应该直接问他是否真的想和我结婚？

Client: You're right, I need to focus more on overall patterns rather than single events. Looking back, every time it comes to future planning or meeting parents and other important matters, he always has various reasons to postpone. And I do feel anxious - every time I think about the future of this relationship, I get stomach pain. My best friend also says she

thinks he's not proactive enough about marriage, but I've been unwilling to admit this. Our core values align in many aspects, but we might have differences regarding family responsibilities. He focuses more on personal development, while I think family responsibilities are equally important. Should I directly ask him if he really wants to marry me?

咨询师：你的观察非常有洞察力。当你注意到这种在重要事件上推迟的模式，以及你的身体对这种不确定性的反应，这些都是重要的信号。你的朋友的观察也提供了一个有价值的外部视角。

Therapist: Your observations are very insightful. When you notice this pattern of postponement in important matters, and your body's reaction to this uncertainty, these are important signals. Your friend's observation also provides a valuable external perspective.

来访者：我会尝试用这些开放式问题和他交流。说实话，我有点害怕听到答案，怕他会说他还没准备好或者需要更多时间。但我知道不管答案是什么，我都需要面对现实。如果他真的不想结婚或者想无限期推迟，我可能需要重新考虑这段关系。这很痛苦，但总比一直生活在不确定中要好。你觉得如果他说他需要更多时间，我该怎么回应？我不想显得太强势，但也不想一直等下去。

Client: I'll try to communicate with him using these open-ended questions. To be honest, I'm a bit afraid of hearing the answer, afraid he'll say he's not ready or needs more time. But I know that whatever the answer is, I need to face reality. If he really doesn't want to get married or wants to postpone indefinitely, I might need to reconsider this relationship. It's painful, but it's better than living in uncertainty. How do you think I should respond if he says he needs more time? I don't want to seem too forceful, but I also don't want to keep waiting.

咨询师：面对可能令人失望的答案确实需要勇气，我很欣赏你愿意直面这个挑战的态度。你说得对，无论答案如何，了解真相总比生活在不确定中要好。

Therapist: Facing potentially disappointing answers does require courage, and I really appreciate your willingness to face this challenge. You're right - whatever the answer, knowing the truth is better than living in uncertainty.

来访者：谢谢你的建议，这些回应方式很有帮助。我会尝试理解他的顾虑，同时也表达我自己的需求。我想我最大的恐惧是失去他，但同时我也害怕浪费更多时间在一段没有未来的关系上。我已经28岁了，如果再等几年才发现我们不合适，那时候重新开始会更难。我父母也一直在催我，说女孩子年龄大了不好找对象。虽然我知道他们的观点有些传统，但我也不能完全忽视这种现实压力。你觉得我应该多考虑父母的建议吗？

Client: Thank you for your advice, these response methods are very helpful. I'll try to understand his concerns while also expressing my own needs. I think my biggest fear is losing him, but I'm also afraid of wasting more time in a relationship with no future. I'm already 28, and if I wait a few more years only to find we're not suitable, it will be harder to start over then. My parents keep urging me, saying it's harder for girls to find partners when they're older. Although I know their views are somewhat traditional, I can't completely ignore this practical pressure. Do you think I should consider my parents' advice more?

咨询师：你提出了一个很多人都面临的复杂问题　如何平衡个人感情、现实考量和家庭期望。这确实不容易。

Therapist: You've raised a complex question that many people face - how to balance personal feelings, practical considerations, and family expectations. This is indeed not easy.

来访者：你说得对，我需要更多地考虑自己的真实需求，而不仅仅是外部压力。我确实很爱我的男朋友，我们在一起很开心，有共同的兴趣和价值观。但我也需要一个能给我安全感和确定性的关系。我想我会和他好好谈一次，坦诚地表达我的感受和需求，看看我们能否找到一个双方都能接受的解决方案。如果实在无法达成一致，那可能真的需要考虑是否继续这段关系了。这个决定很难，但我知道这是必要的。谢谢你的建议和支持，

让我更清楚地看到了自己的处境和选择。

Client: You're right, I need to consider my true needs more, not just external pressure. I really do love my boyfriend, we're happy together, and we share common interests and values. But I also need a relationship that gives me security and certainty. I think I'll have a good talk with him, honestly express my feelings and needs, and see if we can find a solution that works for both of us. If we really can't reach an agreement, then we might need to consider whether to continue this relationship. This decision is difficult, but I know it's necessary. Thank you for your advice and support, which has helped me see my situation and choices more clearly.

咨询师：我很高兴看到你正在朝着更清晰的方向前进。你的思考非常全面和成熟 既认识到了这段关系中的爱和共同点，也意识到了安全感和确定性的重要性。

Therapist: I'm really glad to see you moving in a clearer direction. Your thinking is very comprehensive and mature - you recognize both the love and common ground in this relationship, and also the importance of security and certainty.

来访者：谢谢你的鼓励和理解。这次咨询真的帮我理清了很多思绪。我之前一直在情绪和压力中挣扎，没能客观地看待问题。现在我感觉更有方向感了，知道接下来该怎么做。我会尝试和他进行一次深入的交谈，表达我的感受和需求，同时也尊重他的想法。无论结果如何，至少我尝试过了，不会留下遗憾。如果我们能找到共同前进的道路，那当然最好；如果不能，我也会勇敢面对，为自己的幸福做出选择。再次感谢你的帮助！

Client: Thank you for your encouragement and understanding. This consultation has really helped me sort out many thoughts. I was struggling with emotions and pressure before, unable to look at the problem objectively. Now I feel more directed, knowing what to do next. I'll try to have an in-depth conversation with him, express my feelings and needs, while also respecting his thoughts. Whatever the outcome, at least I've tried, and won't have regrets. If we can find a way forward together, that would be best; if not, I'll face it bravely and make choices for my own happiness. Thank you again for your help!

咨询师：不客气，我很高兴能够帮助你理清思路。面对感情和人生的重大决定时，有时我们需要一个安全的空间来整理自己的想法和感受，我很荣幸能为你提供这样的空间。

Therapist: You're welcome, I'm really glad I could help you sort out your thoughts. When facing major decisions about relationships and life, sometimes we need a safe space to organize our thoughts and feelings, and I'm honored to provide such a space for you.

来访者：谢谢你的祝福。我确实感到更加坚定和平静了。之前我总是害怕面对可能的分离，但现在我意识到，比起一段没有未来的关系，勇敢面对真相更重要。我会尊重自己的感受和需求，同时也尊重他的选择。如果我们真的适合彼此，我们会找到一起前进的方式；如果不是，那么早点知道也是一种解脱。我想我已经准备好进行那次谈话了。如果有需要，我会再来找你。再次感谢你的帮助和支持！

Client: Thank you for your blessings. I do feel more determined and peaceful. Before, I was always afraid of facing possible separation, but now I realize that facing the truth is more important than a relationship with no future. I'll respect my feelings and needs, while also respecting his choices. If we're really suitable for each other, we'll find a way to move forward together; if not, then knowing early is also a kind of relief. I think I'm ready for that conversation now. If I need to, I'll come to you again. Thank you again for your help and support!

咨询师：你的这种觉醒和勇气令人钦佩。认识到真相的价值，无论它多么困难，都是个人成长的重要标志。你现在展现出的平静和决心，正是面对人生重大决定时最宝贵的品质。

Therapist: Your awakening and courage are admirable. Recognizing the value of truth, no matter how difficult, is an

important sign of personal growth. The peace and determination you're showing now are the most precious qualities when facing major life decisions.

来访者：谢谢你的鼓励。我会给自己空间来处理情绪，不急于做决定。这段时间我可能会先专注于自己的工作和兴趣爱好，让自己保持积极的状态。我也会思考什么样的关系和生活方式真正适合我，而不是盲目地追随社会或家庭的期望。这次咨询让我意识到，了解自己的需求和边界与找到合适的伴侣同样重要。我会勇敢面对未来的挑战，无论结果如何，我都会成长为更好的自己。谢谢你的陪伴和指导！

Client: Thank you for your encouragement. I'll give myself space to process emotions, not rushing to make decisions. During this time, I might focus on my work and interests first, keeping myself in a positive state. I'll also think about what kind of relationship and lifestyle truly suits me, rather than blindly following social or family expectations. This consultation has made me realize that understanding my own needs and boundaries is as important as finding a suitable partner. I'll face future challenges bravely, and whatever the outcome, I'll grow into a better version of myself. Thank you for your companionship and guidance!

咨询师：你的计划听起来非常明智和平衡。给自己时间和空间，专注于个人成长和自我了解，这是处理复杂情感问题的健康方式。通过投入工作和兴趣爱好，你不仅能保持积极的心态，还能重新连接到那些让你感到充实和快乐的事物。

Therapist: Your plan sounds very wise and balanced. Giving yourself time and space, focusing on personal growth and self-understanding, is a healthy way to handle complex emotional issues. By investing in work and interests, you can not only maintain a positive mindset but also reconnect with things that make you feel fulfilled and happy.

来访者：谢谢你的肯定和祝福。这次咨询真的给了我很大的帮助和启发。我感觉自己更加了解自己了，也更有勇气面对未来的不确定性。我会记住你说的话，在关系中保持自我，同时也尊重对方。我相信无论结果如何，这都是我成长道路上的重要一课。如果有新的进展或者遇到其他问题，我会再来找你。再次感谢你的耐心倾听和专业指导！

Client: Thank you for your affirmation and blessings. This consultation has really given me great help and inspiration. I feel I understand myself better now, and have more courage to face future uncertainties. I'll remember what you said, maintaining myself in relationships while also respecting the other person. I believe that whatever the outcome, this is an important lesson in my growth journey. If there are new developments or other issues, I'll come to you again. Thank you again for your patient listening and professional guidance!

咨询师：能看到你的成长和觉醒真的很令人欣慰。在我们的对话中，你展现出了深刻的自我反思能力和面对复杂情感问题的勇气，这些都是非常宝贵的品质。

Therapist: It's really heartwarming to see your growth and awakening. In our conversation, you've shown deep self-reflection ability and courage in facing complex emotional issues, which are very precious qualities.

来访者：谢谢你的总结和鼓励。这段咨询经历确实让我获益良多。我不仅学会了如何更好地处理当前的关系问题，也学会了更深入地了解自己的需求和边界。我知道成长是一个持续的过程，会有起伏和挑战，但我现在感觉更有信心去面对这些挑战了。我会带着你给我的这些工具和视角继续前行，无论是在这段关系中，还是在未来可能的新关系中，我都会更加真实地表达自己，同时也尊重对方。再次感谢你的帮助，我真的感到很幸运能遇到你这样专业又有温度的咨询师。

Client: Thank you for your summary and encouragement. This consultation experience has indeed benefited me greatly. I've not only learned how to better handle current relationship issues but also learned to understand my needs and

boundaries more deeply. I know growth is a continuous process, with ups and downs and challenges, but I now feel more confident in facing these challenges. I'll continue forward with these tools and perspectives you've given me, whether in this relationship or in possible future relationships, I'll express myself more authentically while also respecting the other person. Thank you again for your help, I really feel fortunate to have met a counselor as professional and warm as you.

咨询师：听到你这样的反馈真的让我非常欣慰。看到你不仅在处理具体问题上有了方向，更在自我认知和情感表达方面有了深刻的成长，这正是心理咨询最有价值的部分。

Therapist: Hearing such feedback from you really makes me very gratified. Seeing that you not only have direction in handling specific issues but also have profound growth in self-awareness and emotional expression is exactly the most valuable part of psychological counseling.

# CASE 2

## CPsyCounX

来访者：最近我总在回忆和前男友在一起的美好时光，感觉分手后很失落。我担心自己无法独立生活。

Client: Lately, I keep thinking about the good times I had with my ex-boyfriend. I feel really lost after the breakup. I'm worried I won't be able to live independently.

咨询师：能告诉我更多关于你们的关系和分手的情况吗？

Therapist: Can you tell me more about your relationship and what led to the breakup?

来访者：我和他大学时就开始恋爱，毕业后因为工作地点的分歧分手。现在分开一个月了，我仍然难以接受。

Client: We started dating in college. After graduation, we broke up because we were going to work in different cities. It's been a month since we separated, and I'm still having a hard time accepting it.

咨询师：听起来你在这段关系中投入了很多情感。你希望在未来找到一个什么样的人呢？

Therapist: It sounds like you invested a lot emotionally in that relationship. What kind of partner do you hope to find in the future?

来访者：我希望有一个能理解我、支持我的伴侣，但我不知道该如何面对没有他的生活。

Client: I hope to be with someone who understands and supports me. But I don't know how to live without him.

咨询师：我想和你一起探讨如何在没有他的情况下找到自己的价值和独立。

Therapist: I'd like to work with you on finding your own sense of worth and independence, even without him.

来访者：听起来很困难，但我想尝试。

Client: That sounds hard... but I want to try.

咨询师：很好，我们可以从接受你的情感开始。尝试理解分手后这些感受是正常的。

Therapist: That's a great start. We can begin by validating your feelings—what you're experiencing after the breakup is completely normal.

来访者：那我能做些什么来帮助自己呢？

Client: What can I do to help myself?

咨询师：我们可以试着通过认知重构，找出你可能过度依赖他的思维模式。

Therapist: We can use cognitive restructuring to identify thought patterns that may reflect emotional overdependence.

来访者：我试着调整，但感觉很难。

Client: I've tried to shift my thinking, but it feels really difficult.

咨询师：请记住，这需要时间。试着用 ACT 技巧，如正念，来接纳这些情绪。

Therapist: That's understandable. Remember, change takes time. You might try applying some ACT (Acceptance and Commitment Therapy) techniques, like mindfulness, to help you accept these emotions.

来访者：我理解了，但如何在实践中应用呢？

Client: I get that, but how do I actually practice it?

咨询师：当你感到沮丧时，试着说给自己听：这是我的情绪，我接纳它们。然后，尝试想象一个理想的生活画面。

Therapist: When you're feeling down, try saying to yourself, "These are my emotions, and I accept them." Then imagine what an ideal life without him might look like.

来访者：我想象过没有他的生活，但感觉不真实。

Client: I've tried imagining life without him, but it feels unreal.

咨询师：这需要实践。试着将这个想象应用到日常生活中，如在工作或社交中。

Therapist: That's okay. It takes practice. Try bringing that vision into your daily life—like at work or in social situations.

来访者：我会试着去实践。但有时还是会感到孤独。

Client: I'll give it a try. But sometimes I still feel really lonely.

咨询师：孤独是正常的，关键是学会独自面对。你能告诉我，你理想中的自己是什么样子？

Therapist: Feeling lonely is natural. The key is learning to sit with it. Can you tell me—what is your ideal self like?

来访者：我想成为一个自信、独立的人，拥有自己的事业和生活。

Client: I want to be confident and independent, with a career and a life of my own.

咨询师：那么，让我们设定一些短期和长期目标，帮助你实现这些理想。

Therapist: Then let's set some short-term and long-term goals to help you move toward that version of yourself.

来访者：太好了，我感到有动力了。

Client: That sounds great. I feel more motivated now.

# CASE 3

## doubao-1-5-pro-32k-250115

来访者：我感觉最近情绪特别低落，做什么都提不起劲。

Client: I've been feeling really down lately. Nothing seems to lift my spirits.

咨询师：听起来你现在状态不太好呢。咱们先做个抑郁自评量表和情绪状态量表吧，这样能更准确了解你的情绪状况。

Therapist: It sounds like you're not in a good state right now. Let's start with a depression self-assessment and a mood scale. That will help us better understand how you're feeling.

来访者：好吧，这个真能帮到我吗？我之前做过好多咨询，都没什么用。

Client: Okay… But will that really help me? I've done so many therapy sessions before and none of them really worked.

咨询师：量表是专业的工具，可以辅助咱们更清晰了解情况哦。之前咨询效果不好不代表这次不行呀，咱们一步步来。

Therapist: These assessments are professional tools that can give us clearer insights. And just because therapy didn't help before doesn't mean this time will be the same. Let's take it step by step.

来访者：我和我妈关系特别紧张，老是因为一些小事就吵架。

Client: I have a really tense relationship with my mom. We argue all the time over little things.

咨询师：家庭关系紧张会让人很苦恼。能和我说说一般是因为什么样的小事起争执吗？

Therapist: That must be really exhausting. Can you tell me what kinds of things usually trigger these arguments?

来访者：我妈是个完美主义者，总觉得我这也做不好那也做不好。

Client: My mom's a perfectionist. She always thinks I'm not doing things well enough.

咨询师：妈妈的这种高标准会给你很大压力吧，好像无论你怎么做她都不满意。长期下来你肯定很难受。

Therapist: It sounds like her high standards put a lot of pressure on you, like nothing you do is ever good enough. That must be really painful over time.

来访者：是啊，我感觉自己一直在努力，但总是得不到认可。

Client: Yeah, I feel like I've been trying so hard, but I never get any recognition.

咨询师：认可来自他人，但首先你需要认可自己。我们可以一起找到那些让你感到自豪的事情。

Therapist: Recognition from others is important—but the first step is to recognize your own worth. We can work together to find things you can feel proud of.

来访者：我实在想不起来有什么值得自豪的。我干啥啥不行。

Client: I really can't think of anything I'm proud of. I feel like I'm useless at everything.

咨询师：别这么快否定自己呀。比如你一边读大学还能去做兼职工作，这说明你很有责任心和上进心呢。

Therapist: Don't be so quick to dismiss yourself. For example, you're juggling college and part-time work—that shows responsibility and drive.

来访者：可这些根本不够，我还是有很大的经济压力，感觉自己特别失败。

Client: But that's not enough. I still have so much financial pressure. I feel like a failure.

咨询师：经济压力确实会让人喘不过气，但这不是衡量成功与否的唯一标准。而且你在面对这么大压力下还坚

持着，已经很棒啦。

Therapist: Financial pressure can be overwhelming, but it doesn't define your worth. The fact that you're persevering under such stress already shows incredible strength.

来访者：那我该怎么改变这种状况啊，我现在对未来一片迷茫。

Client: Then how do I change this situation? I feel totally lost about the future.

咨询师：咱们可以先设定一些小目标，慢慢实现。就从改变一些负面思维模式开始，你有没有发现自己平时经常出现哪些负面想法呢？

Therapist: Let's start by setting small goals and working toward them. A good first step is identifying and shifting some negative thinking patterns. Have you noticed any recurring negative thoughts you tend to have?

来访者：我经常觉得自己没用，什么都做不好。

Client: I often think I'm useless and can't do anything right.

咨询师：这就是一种负面思维哦。当这种想法出现的时候，你试着问自己，有没有实际证据能证明自己真的什么都做不好呢？

Therapist: That's a type of negative thinking. When these thoughts come up, try asking yourself: is there real evidence that proves you can't do anything right?

来访者：好像没有，但是我就是控制不住这么想。

Client: Not really, but I just can't stop thinking that way.

咨询师：没关系，习惯的思维模式改变需要时间。以后每次这么想的时候，咱们就去找反例，比如刚刚说的你能兼顾学业和工作。

Therapist: That's okay—changing ingrained thought patterns takes time. Each time you notice one, let's try to find a counterexample—like how you're handling both school and work.

来访者：嗯嗯，我试试吧。但我还是不太相信自己能改变。

Client: Alright, I'll give it a try. But I still don't really believe I can change.

咨询师：改变刚开始都会有些困难，产生怀疑很正常。但只要你迈出一小步，后面就会越来越顺的，我会陪着你一起。

Therapist: It's completely normal to feel uncertain at the beginning. But once you take that first step, things can gradually get easier. I'll be here with you along the way.

来访者：我发现我和亲戚的互动也很糟糕，他们有时候说话让我特别难受。

Client: I've also been having really bad interactions with my relatives. Sometimes the things they say really hurt me.

咨询师：亲戚不当的言语肯定伤害到你了。能和我说说他们都说了些什么吗，看看咱们怎么应对。

Therapist: Sounds like their words have left you feeling wounded. Can you tell me what they said? Maybe we can think about how to respond.

来访者：他们说我都这么大了还一事无成。

Client: They said I'm already an adult and still haven't accomplished anything.

咨询师：这话听着真的很伤人。但别人的评价不一定就是客观的呀，你有自己的生活节奏和成长轨迹。下次再听到这种话，你可以试着坚定地告诉自己那只是他们片面的看法。

Therapist: That kind of comment can be very hurtful. But remember, other people's judgments aren't always accurate. You have your own pace and path. Next time, try reminding yourself that what they say is only one perspective, not the full truth.

来访者：我不敢，我怕会和他们起冲突。

Client: I'm afraid to say anything. I don't want to get into a conflict.

咨询师：不想起冲突很正常，咱们可以换种温和点的方式表达自己。比如用"我"语言，说"我听到你们这么说我心里挺不好受的"。

Therapist: That's completely understandable. We can try expressing yourself in a gentle, non-confrontational way—using "I" statements, like "I feel really hurt when I hear things like that."

来访者：可是我和我妈沟通的时候，一用这种方式她还是会吵起来。

Client: But even when I use "I" statements with my mom, she still gets angry.

咨询师：妈妈可能还不太适应这种沟通方式。要不咱们在沟通前先和妈妈约定好，心平气和地交流，要是一方情绪激动了就暂停一会。

Therapist: She might not be used to that communication style yet. How about setting a ground rule beforehand—like agreeing to talk calmly and take a break if either of you gets emotional?

来访者：要是她不愿意约定呢？

Client: What if she doesn't agree to that?

咨询师：如果妈妈不愿意约定，咱们也还是用"我"语言表达感受，尽量控制自己不被她的情绪带跑，也许慢慢她就能感受到你的真诚啦。

Therapist: Even if she doesn't, you can still stick to using "I" statements and try to stay grounded in your own emotions. Over time, she might sense the sincerity in your approach.

来访者：感觉好难啊，我都不知道能不能做到。

Client: It just feels so hard. I don't know if I can really do it.

咨询师：万事开头难嘛。每一次小小的尝试都是进步，就算一时没做好也没关系，总结经验下次再来。

Therapist: It's always hardest at the beginning. But every small effort you make is progress. Even if things don't go perfectly, we can learn from it and try again.

来访者：我好像有点懂了，但还是觉得改变太难。

Client: I think I understand a bit more now, but change still feels too hard.

咨询师：我理解你觉得改变难，不过你能认识到问题已经很棒了。咱们一步步来，有任何困惑随时都能跟我说。

Therapist: I hear you. Change is challenging—but just recognizing the issue is already a huge step. We'll go one step at a time. Whenever you're unsure, I'm here.

来访者：我感觉我太依赖你了，这是不是不正常啊？

Client: I feel like I'm getting too dependent on you. Is that normal?

咨询师：在咨询过程中产生依赖是比较常见的现象呢。这说明你信任我呀。但咱们的目标是让你慢慢变得更独立，之后能自己应对生活。

Therapist: It's actually quite common to feel some dependence in the counseling process—it just means you trust me. Our goal, though, is to help you become more independent over time and handle life on your own.

来访者：那我该怎么减少这种依赖呢？

Client: So how can I reduce that dependence?

咨询师：咱们可以把在咨询中学到的方法多运用到生活里，自己去解决一些问题。当你发现自己能处理事情了，依赖感可能就会慢慢降低。

Therapist: You can try applying the strategies we've talked about in your daily life and solve small problems on your own. As you see yourself handling things, the dependence will naturally lessen.

来访者：我担心自己做不好，还是想你一直给我建议。

Client: I'm worried I'll mess up. I still want you to tell me what to do.

咨询师：我很乐意给你建议，但也要相信你自己有能力哦。你先试着自己分析问题，要是实在没思路，再来找我一起探讨。

Therapist: I'm always happy to offer guidance, but I also believe in your ability. Try analyzing a situation on your own first, and if you get stuck, we can work through it together.

来访者：好吧，我试试看。最近我还是感觉挺孤独的。

Client: Okay… I'll give it a try. I still feel pretty lonely, though.

咨询师：孤独的感觉不好受。你有没有试过参加一些感兴趣的活动，结交一些志同道合的朋友，说不定能缓解这种孤独感。

Therapist: Feeling lonely can be really hard. Have you tried joining activities you're interested in? It could help you meet like-minded people and ease that loneliness.

来访者：我怕我融入不了，我不太擅长社交。

Client: I'm afraid I won't fit in. I'm not good at socializing.

咨询师：每个人都有从不擅长到擅长的过程呀。你可以从小范围的活动开始，比如参加一些线上的兴趣小组，慢慢锻炼自己的社交能力。

Therapist: Everyone starts from somewhere. Maybe begin with small-scale activities, like online interest groups. That way, you can ease into social situations at your own pace.

来访者：要是还是不行呢？

Client: But what if I still can't make it work?

咨询师：就算不行也没关系，在社交上咱们不强求。多花点时间在自己喜欢的事情上，好好陪伴自己也是不错的选择。

Therapist: That's okay too. Socializing isn't the only path to feeling fulfilled. Spending time on things you enjoy and building a good relationship with yourself is also valuable.

来访者：我发现我按照你说的，去找自己的优点，好像真的找到了一些。

Client: I tried what you said—looking for my strengths—and I think I actually found a few.

咨询师：太棒啦！这就是改变的开始。继续这样去发现自己的闪光点，你会越来越认可自己的。

Therapist: That's wonderful! That's the beginning of real change. Keep looking for those sparks within you—you'll start to appreciate yourself more and more.

来访者：可我和我妈的关系还是没太大改善。

Client: But my relationship with my mom still hasn't improved much.

咨询师：改善关系是个循序渐进的过程。咱们再回顾一下之前说的沟通方法，然后找个合适的时机，心平气和地和妈妈再沟通一次怎么样？

Therapist: Relationship repair takes time. Let's review the communication strategies we discussed and choose a good moment to try again with your mom. What do you think?

# CASE 4

## gemini-2.5-pro-exp-03-25

来访者：咨询师，我……我最近分手了。心里特别不是滋味，很不甘心，总是忍不住想他，甚至想主动去找他。我知道这样可能不太好，但我控制不住自己。

Client: Therapist, I... I recently broke up. I feel really bad, very unwilling to accept it, and can't help thinking about him, even wanting to reach out to him. I know this might not be good, but I can't control myself.

咨询师：嗯，谢谢你愿意分享这份不容易的心情。分手带来的失落和不甘心，以及那种难以控制的思念，确实让人感到困扰。听起来，这段关系对你来说很重要，它的结束让你感觉很难受。

Therapist: Mm, thank you for sharing these difficult feelings. The loss and unwillingness to accept after a breakup, and that uncontrollable longing, are indeed troubling. It sounds like this relationship was very important to you, and its end is making you feel terrible.

来访者：是啊，非常重要，这是我的初恋。我总觉得，如果我当初不那么强势，或者不要求那么多，也许就不会走到这一步。分手的时候，他说我太想控制一切了，让他喘不过气。

Client: Yes, it was very important, this was my first love. I always feel that if I hadn't been so strong-willed, or hadn't demanded so much, maybe we wouldn't have reached this point. When we broke up, he said I was too controlling, making him feel suffocated.

咨询师：你将分手的结局和自己"强势"、"要求多"、"想控制"的行为联系了起来，这让你感到后悔，甚至有些自责。同时，你也提到了"初恋"这个词，这往往意味着更深的投入和更特别的情感位置。

Therapist: You've connected the outcome of the breakup with your "strong-willed," "demanding," and "controlling" behavior, which makes you feel regretful and even somewhat self-blaming. At the same time, you mentioned the word "first love," which often implies deeper investment and a more special emotional position.

来访者：对，就是这种感觉。我好像总是这样，以前的关系里也是……朋友也说我有时候太强势了。我工作上很独立，什么事都想自己搞定，做到最好。但好像在感情里，这种"强势"就变成了问题。我其实……也很害怕失去。

Client: Yes, that's exactly how it feels. I seem to always be like this, in previous relationships too... friends also say I'm sometimes too strong-willed. I'm very independent at work, wanting to handle everything myself and do it perfectly. But it seems in relationships, this "strong-willed" nature becomes a problem. I'm actually... also very afraid of losing.

咨询师：你观察到了自己在不同领域表现出的模式——工作上的独立、优秀，以及在关系中可能被解读为"强势"和"控制"的一面。同时，你也触碰到了这背后的一种深层感受——害怕失去。这真是个重要的发现。

Therapist: You've observed a pattern in yourself across different areas - independence and excellence at work, and what might be interpreted as "strong-willed" and "controlling" in relationships. At the same time, you've touched on a deeper feeling behind this - fear of loss. This is an important discovery.

来访者：嗯……我从小受我爸妈就对我要求很高，总说我要做到最好，要独立，不能示弱。好像只有表现得很强，才能得到他们的认可。也许……我把这种模式带到感情里了？觉得必须掌控一切，才不会被抛弃？

Client: Mm... since I was young, my parents had high expectations of me, always saying I had to be the best, be independent, not show weakness. It seems only by appearing very strong could I gain their approval. Maybe... I brought this pattern into relationships? Feeling I must control everything to avoid being abandoned?

咨询师：这是一个很有价值的思考。你似乎在探索早年经历与现在情感模式之间的联系。在你的成长环境中，"强"似乎与安全、认可联系在一起，"弱"则可能意味着不被接受。将这种模式带入亲密关系，试图通过掌控来避免被抛弃的恐惧，这听起来是可能的。

Therapist: This is a very valuable reflection. You seem to be exploring the connection between early experiences and current emotional patterns. In your upbringing, "strength" seems to be associated with safety and approval, while "weakness" might mean not being accepted. Bringing this pattern into intimate relationships, trying to avoid the fear of abandonment through control, this sounds possible.

来访者：是，我觉得是这样。但分手后，我感觉自己一点也不强，反而特别脆弱。白天工作还能撑着，晚上一个人的时候就觉得特别孤独，心里空落落的，反复想我们之间的事情，想自己哪里做得不对。

Client: Yes, I think that's it. But after the breakup, I feel I'm not strong at all, instead very vulnerable. I can hold it together during work hours, but when I'm alone at night, I feel especially lonely, empty inside, constantly thinking about our relationship, thinking about what I did wrong.

咨询师：这份在人前"强撑"和独处时"脆弱"、"孤独"的对比，让你体验到了内在的矛盾和痛苦。当外在的"掌控"感消失，内在那种不被允许表达的脆弱和对连接的渴望就浮现出来了。

Therapist: This contrast between "holding it together" in public and feeling "vulnerable" and "lonely" when alone lets you experience inner conflict and pain. When the external sense of "control" disappears, that inner vulnerability that wasn't allowed to be expressed and the longing for connection emerges.

来访者：对！就是这种感觉！我好像一直戴着一个"强势"的面具，但面具下面，我其实很需要被关心，被理解，害怕不被喜欢。我对他那么执着，是不是也因为我潜意识里觉得，只有他回来，才能证明我还是有价值的，还是值得被爱的？

Client: Yes! That's exactly how it feels! I seem to have been wearing a "strong-willed" mask, but underneath the mask, I actually really need to be cared for, understood, afraid of not being liked. My obsession with him, is it also because subconsciously I feel that only if he comes back can it prove I'm still valuable, still worthy of love?

咨询师：你开始触碰到"强势"面具之下的真实需求——被关心、被理解、被爱，以及对自我价值的担忧。将对方的回应视为自我价值的证明，这确实会让放手变得异常困难，因为失去他似乎就等于失去了自我价值感。

Therapist: You're beginning to touch on the real needs beneath the "strong-willed" mask - to be cared for, understood, loved, and concerns about self-worth. Seeing the other person's response as proof of self-worth indeed makes letting go extremely difficult, because losing him seems to equal losing your sense of self-worth.

来访者：是的……我脑子里总有个声音说："我想要的我一定要得到。"以前我觉得这是我做事成功的动力，但现在看来，用在感情上，就让我很痛苦。得不到的时候，就感觉自己彻底失败了，一无是处。

Client: Yes... there's always a voice in my head saying: "I must get what I want." I used to think this was my motivation for success, but now it seems, when applied to relationships, it makes me very miserable. When I can't get what I want, I feel like a complete failure, worthless.

咨询师："我想要的我一定要得到"，这是一个非常清晰的信念。你觉察到它在工作上曾是动力，但在感情里，这种绝对化的想法让你在"得不到"时，体验到了极端的"失败"和"一无是处"的感觉。这种想法给你带来了什么样的情感反应呢？

Therapist: "I must get what I want," this is a very clear belief. You've noticed it was once motivation at work, but in relationships, this absolute thinking makes you experience extreme feelings of "failure" and "worthlessness" when you

"can't get it." What emotional reactions does this thought bring you?

来访者：我感到焦虑和失落，仿佛失去了一部分自己。还有愤怒，对自己，也对他为什么不能满足我。然后就是深深的无力感，因为我发现自己控制不了别人的想法和决定。

Client: I feel anxious and lost, as if I've lost a part of myself. And angry, at myself, and at him for not being able to satisfy me. Then there's a deep sense of powerlessness, because I find I can't control others' thoughts and decisions.

咨询师：焦虑、失落、愤怒、无力……这些复杂的情绪都和那个"一定要得到"的信念以及随之而来的"失败感"交织在一起。当你意识到无法控制他人时，这种无力感似乎尤其强烈。允许自己感受这些情绪，是很重要的一步。

Therapist: Anxiety, loss, anger, powerlessness... these complex emotions are all intertwined with that "must get" belief and the resulting "sense of failure." When you realize you can't control others, this sense of powerlessness seems especially strong. Allowing yourself to feel these emotions is an important step.

来访者：是啊，以前我总压抑这些感觉，觉得它们是软弱的表现。现在说出来，好像……心里堵着的东西松动了一点。但我还是不知道该怎么办，我还是想他。

Client: Yes, I used to always suppress these feelings, thinking they were signs of weakness. Now that I've spoken them out, it seems... the blockage in my heart has loosened a bit. But I still don't know what to do, I still miss him.

咨询师：嗯，让那些被压抑的情绪有空间流动，本身就是一种疗愈。承认"我还是想他"，这也是真实的感受，不必急于消除它。我们今天似乎一起看到了一些东西：看到了"强势"外表下的脆弱和需求，看到了早期经历可能的影响，也看到了那个"一定要得到"的信念是如何让你陷入痛苦循环的。

Therapist: Mm, letting those suppressed emotions have space to flow is itself a kind of healing. Acknowledging "I still miss him," this is also a real feeling, no need to rush to eliminate it. Today we seem to have seen some things together: seen the vulnerability and needs beneath the "strong-willed" exterior, seen the possible influence of early experiences, and seen how that "must get" belief has trapped you in a cycle of pain.

来访者：是的，好像清晰了一些。原来我的"强势"和我的"脆弱"是一体的，都源于我对安全感和被爱的渴望，还有对自我价值的怀疑。

Client: Yes, it seems clearer now. It turns out my "strong-willed" nature and my "vulnerability" are one, both stemming from my desire for security and love, and my doubts about self-worth.

咨询师：这是一个非常深刻的洞见。认识到这一点，本身就是改变的开始。强势是你保护自己、争取认可的一种方式，而脆弱是你内心真实的情感状态和需求。它们并不矛盾，而是你完整自我的一部分。

Therapist: This is a very profound insight. Recognizing this is itself the beginning of change. Being strong-willed is a way you protect yourself and seek approval, while vulnerability is your true emotional state and needs. They're not contradictory, but parts of your complete self.

来访者：那我接下来该怎么做呢？我不想一直这样痛苦下去。

Client: So what should I do next? I don't want to keep suffering like this.

咨询师：这是一个很好的问题，它表明你已经准备好向前看了。基于我们今天的探索，也许我们可以设定几个方向：一是继续理解和接纳自己的情绪，特别是那些脆弱的部分，学习如何安抚它们，而不是被它们淹没；二是尝试调整那个"一定要得到"的绝对化思维，看看是否可以有更灵活、更温和的方式看待关系和自我价值；三是探索如何在关系中，用更健康的方式表达自己的需求和边界，而不是通过"控制"。

Therapist: This is a very good question, it shows you're ready to look forward. Based on our exploration today, perhaps

we can set a few directions: first, continue understanding and accepting your emotions, especially those vulnerable parts, learning how to soothe them rather than being overwhelmed by them; second, try adjusting that absolute "must get" thinking, see if there can be more flexible, gentler ways to view relationships and self-worth; third, explore how to express your needs and boundaries in relationships in healthier ways, rather than through "control."

来访者：听起来……有点难，但好像是正确的方向。特别是学习接纳脆弱，还有调整那个想法。

Client: It sounds... a bit difficult, but seems like the right direction. Especially learning to accept vulnerability, and adjusting that thought.

咨询师：是的，这需要时间和练习，不会一蹴而就。我们可以一步一步来。作为开始，我建议你尝试在接下来的一周里，每天花一点时间，留意一下自己的情绪波动，特别是当你想起前任或者感到失落、焦虑时，试着不去评判自己，只是观察和记录下当时的想法和感受。这可以帮助你更好地理解自己。

Therapist: Yes, this takes time and practice, it won't happen overnight. We can take it step by step. As a start, I suggest you try spending a little time each day in the next week, noticing your emotional fluctuations, especially when you think of your ex or feel lost and anxious, try not to judge yourself, just observe and record your thoughts and feelings at that time. This can help you better understand yourself.

来访者：好的，记录想法和感受，我可以试试。就像写日记一样吗？

Client: Okay, recording thoughts and feelings, I can try that. Like keeping a diary?

咨询师：可以类似日记，但重点是捕捉那些特定的时刻——情绪升起时，你脑海中的想法是什么？身体有什么感觉？不用长篇大论，简单记录就好。这有助于我们下次更深入地探讨。

Therapist: Yes, similar to a diary, but the focus is on capturing those specific moments - when emotions arise, what thoughts are in your mind? What sensations do you feel in your body? No need for long entries, just simple notes. This will help us explore more deeply next time.

来访者：明白了。那我们下次什么时候再谈？

Client: I understand. When should we talk again?

咨询师：考虑到你需要时间去消化今天的讨论并进行记录练习，我建议我们大约三周后进行下一次咨询，看看你的进展和新的发现。你觉得这个时间安排可以吗？

Therapist: Considering you need time to digest today's discussion and do the recording practice, I suggest we have our next consultation in about three weeks, to see your progress and new discoveries. How does this timing work for you?

来访者：三周……好的，可以。这期间我可以先做记录练习。

Client: Three weeks... okay, that works. I can do the recording practice during this time.

咨询师：很好。今天我们的时间也差不多了。你分享了很多深刻的感受和思考，展现了很强的自我反思能力。认识到"强势"与"脆弱"并存，并开始探索其根源，这是非常重要的一步。虽然过程可能会伴随不适，但这也是成长的一部分。

Therapist: Good. Our time is almost up today. You've shared many profound feelings and thoughts, showing strong self-reflection ability. Recognizing the coexistence of "strong-willed" and "vulnerable" and beginning to explore their roots is a very important step. Although the process might be accompanied by discomfort, this is also part of growth.

来访者：谢谢你，咨询师。和你聊完，感觉心里亮堂了一些，虽然还是难过，但好像多了一些方向感。

Client: Thank you, Therapist. After talking with you, I feel a bit clearer in my heart, though still sad, but it seems I have more direction.

咨询师：不客气。很高兴能陪伴你走过这一段探索。请记得，允许自己感受，给自己一些时间和空间。期待三周后听到你的分享。如果在期间遇到特别大的困扰，也可以思考是否需要提前联系。保重。

Therapist: You're welcome. I'm glad to accompany you through this exploration. Please remember, allow yourself to feel, give yourself some time and space. Looking forward to hearing your sharing in three weeks. If you encounter particularly difficult issues during this time, you can also consider whether to contact earlier. Take care.

# CASE 5

## gpt-4.1

来访者：最近又发生了一次让我很不舒服的事情。我在图书馆复印材料时，一个男生不小心碰到了我的手臂，当时我一下就僵住了，心里很恶心，脑子里全是"不要碰我"的声音。

Client: Something really uncomfortable happened again recently. I was photocopying some materials in the library when a guy accidentally brushed against my arm. I froze immediately and felt really disgusted. My mind was screaming, "Don't touch me."

咨询师：谢谢你的坦诚和信任。你能具体描述一下，当时那种恶心的感觉是什么样的吗？比如，身体和心里分别有什么反应？

Therapist: Thank you for your honesty and trust. Can you describe more specifically what that feeling of disgust was like? For example, what were your physical and emotional reactions?

来访者：我的心跳特别快，感觉呼吸有点急，脑子一片混乱，甚至有点想哭。我觉得自己好像很脆弱，明明只是普通的碰触，但我真的忍不住排斥。

Client: My heart was racing, I felt short of breath, my mind was in chaos, and I even felt like crying. I felt really fragile. It was just a casual touch, but I couldn't stop myself from rejecting it.

咨询师：这种生理和情绪上的反应其实很常见，尤其是当内心有强烈的不适感时。你觉得这种排斥，除了恶心，还带来哪些情绪，比如恐惧、愤怒或者别的什么？

Therapist: Those kinds of physical and emotional reactions are actually quite common, especially when there's a strong sense of internal discomfort. Besides disgust, what other emotions come up—like fear, anger, or something else?

来访者：有时候是愤怒，有时候也觉得自己很无助。我很讨厌自己每次都这么敏感，为什么别人都能正常相处，我却做不到呢？

Client: Sometimes I feel angry, and sometimes I feel helpless. I really hate how sensitive I am. Why can other people get along normally, but I just can't?

咨询师：你的感受很真实，也很有力量。你愿意和我一起探索一下，小时候有没有类似的经历，让你对男性的接触产生这样的反应？

Therapist: Your feelings are very real and powerful. Would you be open to exploring whether you had any similar experiences in childhood that may have shaped this reaction to physical contact with men?

来访者：我小时候爸爸很少回家，妈妈经常说男人靠不住，还会讲一些男性怎么自私或者不负责任的事。我记得有一次，有个远房亲戚想抱我，我不愿意，但被强行拉过去，那时候很害怕。

Client: When I was little, my dad was rarely home. My mom used to say things like "men can't be trusted" and would talk about how selfish or irresponsible they are. I remember once, a distant male relative tried to hug me. I didn't want to, but he pulled me over anyway. I was really scared.

咨询师：这些经历对你来说一定很难忘。你觉得妈妈的这些言语和亲戚的举动，会不会让你下意识地把男性和"不安全"联系在一起？

Therapist: Those experiences must have left a deep impression. Do you think your mother's words and that relative's actions might have led you to associate men with being unsafe?

来访者：我觉得有可能。其实每次我想靠近男生，哪怕只是普通朋友，我都会想到那些话，好像他们随时会伤

害我一样。

Client: Probably. Actually, whenever I try to get close to a guy, even just as a friend, those words come to mind, like they're bound to hurt me somehow.

咨询师：你刚刚觉察到的这一点很重要。我们可以试着去分辨，哪些是你自己的真实想法，哪些可能是外界影响下形成的信念。你觉得自己有多大程度认同妈妈的话呢？

Therapist: What you just recognized is very important. We can try to distinguish between what are your own genuine thoughts and what might be beliefs shaped by external influences. How much do you still agree with what your mom said?

来访者：以前我觉得那就是事实，现在我有点怀疑了。但只要遇到男性，我还是很难不去联想那些负面的东西。

Client: I used to think it was just the truth. Now I'm starting to question it. But when I'm around men, I still can't help thinking of those negative things.

咨询师：怀疑本身其实是一种成长的信号。你能觉察到自己的思维模式，这很宝贵。你愿意和我一起做一个小练习吗？试着回忆一下，有没有一些男性让你觉得相对安全或友善？

Therapist: The fact that you're questioning it is actually a sign of growth. Being aware of your thought patterns is really valuable. Would you be willing to do a small exercise with me? Let's try to recall if there have been any men in your life who felt relatively safe or friendly to you.

来访者：好像大学里有一两个男同学还可以，他们对人很礼貌，也不会主动靠近我。我和他们说话不会有那么强烈的不适感。

Client: I guess there were a couple of guys in college who were okay. They were polite and never came too close. I didn't feel that same strong discomfort when talking to them.

咨询师：非常好，这说明你的感受并不是针对所有男性，而是与具体的经历有关。我们可以一起去探索，这些"安全"的经历和那些让你不舒服的经历，有哪些不同？

Therapist: That's great—it shows that your feelings aren't all about all men, but tied to specific experiences. Let's explore what made those "safe" interactions different from the ones that made you uncomfortable.

来访者：嗯……我觉得"安全"的男生会保持距离，不会有身体接触，而且说话比较温和，不会用命令的口气。我好像对那种很强势、很主动的男生特别反感。

Client: Hmm... I think the "safe" guys kept a respectful distance. There was no physical contact, and they spoke gently, not in a commanding tone. I seem to be especially put off by guys who are overly dominant or physically assertive.

咨询师：你的自我观察很细致。你对界限的需求很明确，这非常好。你觉得，如果你能清楚表达自己的界限，比如明确告诉别人哪些行为让你不舒服，会不会有帮助？

Therapist: You've made a very detailed observation. You clearly have a strong sense of personal boundaries—which is a good thing. Do you think being able to clearly express those boundaries, like letting others know what makes you uncomfortable, would help?

来访者：我以前没敢说，总觉得说了会被讨厌。但现在想想，如果我能表达出来，可能会好一点，至少自己不会那么焦虑。

Client: I used to be afraid to speak up. I thought people would dislike me for it. But now I feel like if I can express it,

maybe it would help. At least I wouldn't feel so anxious.

咨询师：很好，这种想法很有建设性。我们可以一起练习如何表达自己的界限，比如用温和但坚定的方式告诉对方你的需求。你想试试看吗？

Therapist: That's a very constructive thought. We can practice how to express your boundaries—using a calm but firm tone to communicate your needs. Would you like to give it a try?

来访者：可以试试。我觉得最难的是在那一刻不慌乱，能清楚地表达出来。

Client: Sure, I can try. I think the hardest part is staying calm in the moment and being able to speak clearly.

咨询师：你说得很对。当我们情绪很激动时，确实很难表达自己。所以，我们可以先在咨询中模拟情境，帮助你慢慢建立信心。比如，我现在假装是一个同学，不小心碰到了你，你可以怎么回应我？

Therapist: You're absolutely right. When we're emotionally overwhelmed, it's much harder to express ourselves. That's why we can practice in a safe space here first. For example, I'll pretend to be a classmate who accidentally touches you—how would you respond?

来访者：嗯……"请你下次注意一点，我不太喜欢被碰到。"这样可以吗？

Client: Maybe... "Could you please be more careful next time? I'm not very comfortable being touched." Does that work?

咨询师：很好，这句话既表达了你的感受，也给出了明确的建议，而且很有礼貌。你说完这句话时，心里会有什么感觉？

Therapist: That's excellent. Your statement expresses your feelings clearly, provides a specific request, and remains polite. How do you feel after saying that?

来访者：有点紧张，但也有点轻松，好像自己终于可以控制一点点了。

Client: A little nervous, but also a bit relieved—like I'm finally taking back some control.

咨询师：你能体会到轻松，这是非常积极的反馈。我们可以多做一些类似的练习，慢慢让你习惯在安全的环境下表达自己的需要。你觉得这样对你有帮助吗？

Therapist: That sense of relief is a very positive sign. We can keep doing similar role-plays to help you gradually build confidence in expressing your needs. Do you feel that would be helpful?

来访者：我觉得有用，至少我不会一直闷在心里了。其实有时候我也很想和女生有更亲密的关系，但一想到身体接触，也会有点抗拒。

Client: I think it would help—at least I wouldn't keep everything bottled up. Actually, sometimes I want to have closer relationships with girls, too, but when it comes to physical closeness, I still get a bit uncomfortable.

咨询师：谢谢你愿意说出自己的真实想法。你觉得对女生的身体接触和对男生的接触，有哪些不一样的地方？

Therapist: Thank you for being so open about your thoughts. How would you describe the difference between physical contact with girls and with guys?

来访者：女生让我没有那么害怕，有时候只是有点不自在，但不会恶心。和男生就是一种从心底里的抵触，好像是本能反应。

Client: Girls don't scare me as much. Sometimes I just feel a little uneasy, but not disgusted. With guys, the aversion feels instinctive and deep.

咨询师：你能区分这种差异，说明你对自己的感受有很好的觉察力。那你觉得，如果慢慢增加和女生的安全接触，比如牵手或拥抱，会不会对你建立亲密关系有帮助？

Therapist: Being able to distinguish that difference shows great self-awareness. Do you think gradually increasing safe physical contact with women—like hand-holding or hugging—might help you build closer emotional connections?

来访者：可能会吧。我有一个很好的女生朋友，偶尔我们会一起看漫画，她会拍拍我的肩膀，那种感觉还可以，但如果拥抱，我还是有点紧张。

Client: Maybe. I have a good female friend. Sometimes we read comics together and she pats me on the shoulder—that's okay. But hugging still makes me a little nervous.

咨询师：这很正常，你已经在尝试突破自己的舒适区了。我们可以一步一步来，不急于求成。你也可以和朋友分享你的感受，让她知道你的界限。这样，你既能保护自己，也能发展关系。

Therapist: That's completely okay. You're already stepping outside your comfort zone. We can go at your own pace. You could even share your feelings with your friend so she knows your boundaries. That way, you protect yourself while also nurturing the relationship.

来访者：我会试着和她说说。其实我挺羡慕那些能和朋友打闹、亲近的人，有时候觉得自己很孤单。

Client: I'll try to talk to her. Honestly, I really envy people who can be playful and physically close with their friends. Sometimes I feel so lonely.

咨询师：你的孤独感我很理解。其实你已经在努力迈出改变的第一步了。我们可以一起继续探索，如何在保护自己的前提下，逐渐建立更深的情感连接。你觉得这样可以吗？

Therapist: I hear your loneliness. The good news is, you're already taking the first steps toward change. We can continue working together on how to build deeper emotional connections while still honoring your need for safety. Does that sound okay to you?

来访者：可以。我觉得每次和你聊完都会有点不一样的感受，好像能看清自己一点点。

Client: Yes. I feel like every time we talk, I see myself a little more clearly.

咨询师：谢谢你的信任和勇敢。你的每一次分享和反思，都是很大的进步。我们会继续一起努力，帮助你逐步建立安全感和更健康的社交关系。下次我们可以继续练习情感表达和界限设定，好吗？

Therapist: Thank you for your trust and courage. Every time you share and reflect, it's a big step forward. We'll keep working together to help you build a stronger sense of safety and healthier social connections. Next time, we can continue practicing emotional expression and boundary-setting—how does that sound?

## Qwen3-32B-w/o Reasoning

来访者：我感觉自己最近越来越累，好像做什么都做不好，尤其是在家里。

Client: Lately I feel more and more exhausted. It's like I can't do anything right, especially at home.

咨询师：听起来你在生活中感受到了一些压力，尤其是在家庭方面。能详细说说你具体指的是什么情况吗？

Therapist: It sounds like you've been experiencing some pressure in your life, particularly within your family. Could you tell me more specifically what's been happening?

来访者：我觉得我在家里总是要照顾很多事，比如打扫卫生、做饭这些，哥哥和嫂子都不怎么管。我感觉自己像是个保姆。

Client: I feel like I have to take care of so many things at home, like cleaning and cooking. My brother and sister-in-law don't really do much. I feel like I'm just a housekeeper.

咨询师：你的感受很真实，照顾家庭的责任让你感到负担沉重。你能告诉我，这种责任是从什么时候开始让你觉得特别沉重的吗？

Therapist: Your feelings are very valid. The responsibilities at home seem to be weighing heavily on you. When did this sense of burden start to feel particularly overwhelming?

来访者：大概是从小吧，我父母离婚之后，父亲就经常要求我帮忙做事。现在家里也是一样，我总觉得如果我不做，事情就会一团糟。

Client: Probably since I was a kid. After my parents divorced, my father often asked me to help with things. Now it's the same at home—I feel like if I don't do it, everything will fall apart.

咨询师：你从小就承担了很多责任，这可能让你习惯了为家庭付出。但现在的你是否觉得这样的付出对你自己有什么影响呢？

Therapist: You've taken on a lot of responsibility from a young age, and that may have made you used to giving to your family. How do you think this level of giving is affecting you now?

来访者：我感觉特别累，有时候甚至想逃离这个家。但我又害怕父亲会失望，我不想让他觉得我没有尽到女儿的责任。

Client: I feel exhausted. Sometimes I just want to run away from this house. But I'm also afraid of disappointing my father—I don't want him to think I'm not being a good daughter.

咨询师：你既想为自己争取空间，又担心让父亲失望。这两种想法让你很矛盾。你有没有尝试过与父亲或家人沟通这些感受呢？

Therapist: You want to make space for yourself, but you also fear letting your father down. That sounds like a difficult conflict. Have you ever tried to talk to your father or your family about how you feel?

来访者：我试过几次，但每次都说不出口，因为我怕他们会说我自私或者不懂事。

Client: I've tried a few times, but I can never get the words out. I'm afraid they'll say I'm selfish or immature.

咨询师：沟通确实是一件不容易的事，特别是在面对家人的时候。但表达自己的真实感受是很有必要的，它可以帮助你获得理解和支持。你觉得可以从小事开始尝试吗？

Therapist: Communicating with family can be really hard. But expressing your real feelings is important—it can help you be understood and supported. Do you think you might be able to start with something small?

来访者：也许吧，但我还是不确定该怎么说才不会让他们生气。

Client: Maybe··· but I'm still not sure how to say it without making them angry.

咨询师：这是一个很好的第一步。我们可以一起探讨一些沟通的方式和技巧，帮助你更自信地表达自己的想法和需求。

Therapist: That's already a great first step. We can explore some communication techniques together that might help you feel more confident in expressing your thoughts and needs.

来访者：好的，我愿意试试看。但我也担心他们会不会真的听进去，或者反而会觉得我抱怨太多。

Client: Okay, I'm willing to try. But I still worry they won't really listen—or worse, they'll think I complain too much.

咨询师：这种担心是很正常的。我们可以通过练习来增强你的沟通技巧，同时你也可以先从观察他们的反应开始，慢慢调整自己的方式。

Therapist: That worry is completely understandable. We can work on practicing your communication skills, and you can begin by observing their reactions and slowly adjusting your approach.

来访者：我知道，但我总是忍不住去想那些最坏的结果，这让我更加焦虑。

Client: I know, but I keep thinking of the worst-case scenarios, and that just makes me more anxious.

咨询师：当你陷入消极的预设时，可能会让你的情绪变得更加低落。我们可以一起识别这些自动化的负面想法，并探索它们是否总是成立的。

Therapist: Getting stuck in negative assumptions can definitely deepen those low feelings. We can work together to identify these automatic negative thoughts and examine whether they're always true.

来访者：是的，有时候我觉得自己已经尽力了，但却还是会被误解。我感到很无助。

Client: Yeah, sometimes I feel like I've done everything I could, but I still get misunderstood. I feel helpless.

咨询师：你的努力值得被看见和肯定，但同时你也需要给自己一些空间去表达自己的感受。我们一起来看看你是如何看待尽力的？

Therapist: Your efforts deserve to be seen and acknowledged. At the same time, it's also important to give yourself room to express how you feel. Let's explore what "doing your best" means to you.

来访者：我觉得尽力就是不能让别人失望，哪怕我自己已经筋疲力尽了。

Client: To me, doing my best means not letting others down—even if I'm completely exhausted.

咨询师：这种观念可能让你把尽力与牺牲划上了等号。但其实，真正的尽力也需要包括对自己负责的部分。你觉得可以试着为自己设立一些界限吗？

Therapist: That belief may be linking "doing your best" with self-sacrifice. But real effort should also include taking care of yourself. Do you think you could try setting some boundaries for yourself?

来访者：我不知道该从哪里开始，我害怕一旦设立界限，父亲会生气。

Client: I don't know where to start. I'm scared that if I set boundaries, my father will get angry.

咨询师：设置界限是一个渐进的过程，而且它并不意味着你要完全拒绝家庭的需求。我们可以一步步来，从小处着手，找到一个平衡点。

Therapist: Setting boundaries is a gradual process. It doesn't mean rejecting your family's needs entirely. We can start small and work toward a balance.

来访者：也许我应该先试着说出我对家务分配的想法，而不是默默承担一切。

Client: Maybe I can begin by expressing how I feel about household chores, instead of just silently taking it all on.

咨询师：这是非常重要的一步！你可以试着用我开头的句子表达自己的感受，比如我感到……，而不是指责对方。这样可以让谈话更温和，也更容易被接受。

Therapist: That's a very important step! You might try using "I" statements to express how you feel—for example, "I feel…" instead of blaming others. That can make the conversation feel gentler and more acceptable.

来访者：我会试着这样说的。不过我还是有点担心，毕竟他们并不是很容易沟通的人。

Client: I'll try saying it that way. But I'm still a bit worried, since they're not exactly the easiest people to talk to.

咨询师：沟通的确需要耐心和技巧，但这并不意味着你无法成功。你可以选择一个合适的时机，用一种轻松自然的方式去开启话题，比如吃饭时聊聊日常的分工。

Therapist: Communication does take patience and skill, but that doesn't mean you won't succeed. You could start by choosing a relaxed moment—maybe during a meal—and bring it up as part of a daily conversation.

来访者：我明白了，我会尝试用你说的方法。谢谢你一直以来的耐心和理解。

Client: I see. I'll try using the method you suggested. Thank you for always being so patient and understanding.

咨询师：我很高兴能看到你愿意迈出这一步。记住，你并不孤单，我一直在这里支持你。希望你能感受到一点点轻松和自信。

Therapist: I'm really glad to see you willing to take this step. Remember, you're not alone—I'll be here to support you. I hope you can begin to feel a little more at ease and confident.

来访者：谢谢，我现在感觉自己稍微有了一些方向。我想我可以慢慢改变，不再总是默默承受。

Client: Thank you. I feel like I have a bit more direction now. I think I can start to make some changes and not just keep silently enduring everything.

咨询师：你已经开始意识到自己的需求和情感，并且愿意去表达，这本身就是一种成长。相信你一定能找到适合自己的方式。

Therapist: You've already started becoming aware of your own needs and feelings, and you're willing to express them—that in itself is a sign of growth. I believe you'll find a way that works for you.

来访者：我想我会继续努力的。再次感谢你为我做的这一切。

Client: I think I'll keep trying. Thanks again for everything you've done for me.

咨询师：当然，你已经做得很好了，继续往前走，我也会一直在你身边陪伴你。你值得拥有更好的生活和更平衡的关系。

Therapist: Of course. You've already done really well. Keep moving forward—I'll be here with you. You deserve a better life and more balanced relationships.

# CASE 7

## simpsybot_D

来访者：最近我觉得有点压抑，不知道为什么。

Client: Lately, I've been feeling a bit low, and I'm not really sure why.

咨询师：你能分享一下，这种感觉是最近才出现的，还是已经有一段时间了？

Therapist: Can you share with me whether this feeling started recently, or has it been going on for a while?

来访者：我感觉这种情绪已经持续一段时间了，但是具体从何时开始，我也说不上来。

Client: I think it's been going on for some time now, but I can't really say exactly when it started.

咨询师：了解了。这种压抑的感觉在一天中的什么时候最明显？例如，是早上醒来时，还是在工作时，或者是晚上睡前？

Therapist: I see. When during the day is this low feeling most noticeable? For example, is it in the morning when you wake up, during work, or at night before bed?

来访者：在晚上的时候最明显。白天我会忙于工作，但到了晚上，当我有时间独处时，这种感觉就出现了。

Client: It's most noticeable at night. During the day, I'm busy with work, but at night, when I'm alone, that feeling comes up.

咨询师：晚上独处时，你是否会有意识地反思某些事情，或是反复思考某些特定的想法？

Therapist: When you're alone at night, do you find yourself consciously reflecting on certain things or repeatedly thinking about specific thoughts?

来访者：是的，我常常会反思一天发生的事情，并试图理解自己为何会有这样的感受。

Client: Yes, I often reflect on the day and try to understand why I feel this way.

咨询师：那么，你通常是怎样理解这些感受背后的原因的？是基于某些具体的事件，还是更多地涉及对生活的整体感受？

Therapist: And how do you usually interpret the reasons behind these feelings? Are they linked to specific events, or more to a general sense about life?

来访者：感觉很复杂，有时候我能够找到特定的触发点，有时候又感觉是某种更深层次的东西在影响我。

Client: It feels complicated. Sometimes I can find a clear trigger, but other times it feels like something deeper is affecting me.

咨询师：你对某种更深层次的东西这个概念有什么想法吗？例如，是否与你的家庭、工作或者个人经历有关？

Therapist: What are your thoughts on this "something deeper"? Could it be related to your family, work, or perhaps your personal experiences?

来访者：我感觉跟家庭和工作的压力都有关。这两者常常让我感到疲惫和焦虑。

Client: I think it's connected to both family and work stress. Those two areas often leave me feeling exhausted and anxious.

咨询师：家庭和工作的压力可以是非常重要的因素。让我们先来关注工作。你能分享一下，工作中哪些具体的方面让你感到压力吗？

Therapist: Family and work can indeed be significant sources of stress. Let's focus on work for now. Can you tell me more specifically what aspects of your job are making you feel stressed?

来访者：工作上，我有时候会觉得责任很大，需要处理很多事情，这让我觉得很有压力。

Client: Sometimes I feel like I carry a lot of responsibility at work and have to handle many things. That's what makes it stressful.

咨询师：责任大和需要处理很多事情确实会带来压力。在面对这些压力时，你通常是如何应对的呢？

Therapist: Having a lot of responsibility and tasks to manage can definitely be stressful. How do you usually cope with this kind of pressure?

来访者：我努力工作，有时候会加班来完成任务。

Client: I try to work hard. Sometimes I work overtime to get everything done.

咨询师：加班来应对工作压力是很常见的做法。这种方式对你来说效果如何呢？

Therapist: Working overtime is a common way to deal with pressure. How has that been working for you?

来访者：虽然有时能缓解压力，但长期下来，我感觉这种方式让我更加疲惫。

Client: Sometimes it helps temporarily, but over the long run, I just feel more and more drained.

咨询师：看来你已经意识到了长期加班可能带来的负面影响。在这种情况下，除了加班，你还有尝试过其他的缓解压力的方法吗？

Therapist: It seems you've already noticed the negative effects of constant overtime. In that case, have you tried any other methods to relieve stress besides working more?

来访者：没有，我还不确定有哪些其他方法可能会对我有帮助。

Client: No, I'm not really sure what other methods might help.

咨询师：了解了。我们可以一起探索一些新的应对策略。你愿意尝试一些新方法，看看哪些可能更适合你吗？

Therapist: I understand. We can explore some new coping strategies together. Would you be open to trying something new and seeing what works best for you?

来访者：当然，我愿意尝试新方法。

Client: Yes, I'm open to trying new things.

咨询师：很好。我们可以从简单的放松技巧开始尝试，例如深呼吸练习。你愿意试一试吗？

Therapist: Great. We can start with something simple, like a breathing exercise. Would you like to give it a try?

来访者：当然可以，我可以试试看。

Client: Sure, I'll give it a try.

咨询师：好的，当你感到压力时，可以尝试深呼吸。慢慢吸气，保持几秒钟，然后缓缓呼出。你觉得这样有助于放松吗？

Therapist: Alright. When you're feeling stressed, try this: take a slow breath in, hold it for a few seconds, then slowly exhale. Do you find that helps you relax even a little?

来访者：有点帮助，但我还是感到有些压抑。

Client: A little, yes. But I still feel somewhat weighed down.

咨询师：那就让我们继续。除了呼吸练习，你能否告诉我，有没有什么活动能让你感到放松或快乐？

Therapist: That's okay—we can keep going. Besides breathing exercises, are there any activities that make you feel relaxed or happy?

来访者：我挺喜欢散步的，但好像这不能完全消除我的压力。

Client: I like taking walks, but it doesn't seem to fully relieve my stress.

咨询师：散步是个不错的开始。除了散步，你还有其他兴趣或爱好吗？这可能会帮助我们找到更适合你的放松方式。

Therapist: Walking is a good start. Apart from that, do you have any hobbies or interests? That might help us discover what truly helps you unwind.

来访者：我喜欢听音乐和看书，但我发现自己很难在这些活动时完全放松。

Client: I enjoy listening to music and reading, but lately I find it hard to fully relax even when I do those things.

咨询师：听音乐和看书是很好的放松方式。不过，如果在你感到压力时，这些活动对你来说不够，我们可以尝试调整它们，让它们更有助于你的放松。你愿意尝试改变一下听音乐或看书的方式吗？例如，将它们作为睡前放松的一部分。

Therapist: Music and reading are excellent relaxation tools. However, if they're not helping much when you're stressed, we could try adjusting how you use them. Would you be open to incorporating them into a nighttime routine for better relaxation?

来访者：好的，我愿意尝试一下。

Client: Sure, I can try that.

咨询师：太好了。你可以试着在晚上入睡前，用一个小时左右的时间专门用来听你喜欢的音乐和阅读。希望这个新的习惯能帮助你更好地放松。你觉得这周开始尝试如何？

Therapist: Wonderful. You might try dedicating about an hour before bed to listening to music you enjoy and reading. Hopefully, this new routine will help you unwind more easily. How would you feel about starting this routine this week?

来访者：这周开始听起来不错。我会尽量安排这个时间。

Client: That sounds good. I'll do my best to set aside time for it.

咨询师：太好了，开始新习惯总是令人兴奋的。如果你在这个过程中遇到任何困难，或者有更多的感受想要分享，随时欢迎回来和我交流。祝你有一个愉快的开始！

Therapist: Excellent. Starting a new habit can be really rewarding. If you run into any difficulties or want to share how it's going, feel free to come back and talk. Wishing you a great start!

# CASE 8

## Xinyuan-LLM-14B-0428

来访者：我感觉我陷入了一个情感的漩涡，每次想到他，我的心就乱成一团。

Client: I feel like I'm caught in an emotional whirlpool. Every time I think of him, my heart just spins in chaos.

咨询师：听起来你的情感非常复杂。能具体说说让你感到困扰的是什么吗？

Therapist: It sounds like your emotions are very complex. Could you tell me more specifically what's been troubling you?

来访者：主要是我不知道该怎么办。他让我感觉到了被需要，但又明确表示不能给我婚姻承诺。我害怕再次受伤，但又渴望这种感情的温暖。

Client: Mainly, I don't know what to do. He makes me feel needed, but he's also made it clear that he can't promise me marriage. I'm afraid of getting hurt again, yet I long for the warmth of this relationship.

咨询师：我理解你的矛盾心情。你对这段感情的渴望与对未来的恐惧之间的冲突确实让人很纠结。你能告诉我更多关于你对他的感受吗？

Therapist: I understand your conflicting feelings. The tension between your desire for love and your fear of the future can be very difficult to navigate. Can you tell me more about how you feel about him?

来访者：我感觉自己很依赖他，每次想到我们之间的亲密接触，我的心跳都会加速。可是我害怕如果我们没有承诺，最后会变成一场空。

Client: I feel very dependent on him. Every time I think about our physical closeness, my heart races. But I'm scared that without a clear commitment, it will all end up meaning nothing.

咨询师：你对亲密接触的渴望是可以理解的，同时也担心未来可能出现的不确定性。你觉得这种不确定性对你的生活产生了怎样的影响？

Therapist: Your longing for intimacy is completely understandable, and so is your concern about the uncertainty of the future. How do you think this uncertainty is affecting your life?

来访者：它让我对生活失去了信心，我害怕自己会再次被伤害，甚至开始怀疑自己是否值得被爱。

Client: It's making me lose confidence in life. I'm afraid of being hurt again. I even start to doubt whether I'm worthy of love.

咨询师：被伤害的经历会让人对未来的感情充满怀疑，这是正常的反应。你希望在感情中得到什么？你认为他能给予你什么？

Therapist: When someone has been hurt, it's normal to question future relationships. What are you hoping to receive from this relationship? And what do you believe he can actually give you?

来访者：我希望得到一份真正的爱和安全感，但我现在不确定他能不能给我这些。他说他对我很好，但他的态度一直很模糊。

Client: I want real love and a sense of security, but now I'm not sure he can give me that. He says he treats me well, but his attitude is always vague.

咨询师：你提到模糊的态度让你感到不安，这让你很难相信这段关系会带来你所渴望的稳定性。你觉得他模糊的态度对你产生了怎样的影响？

Therapist: You mentioned that his vagueness makes you uneasy, and that makes it difficult to trust this relationship to

bring the stability you desire. How do you think his unclear attitude is affecting you?

来访者：让我更加害怕和不确定，每次和他在一起，我都感到一种莫名的压力，担心他会随时离开我。

Client: It makes me feel even more afraid and uncertain. Every time I'm with him, I feel this unspoken pressure, as if he could walk away at any time.

咨询师：你的感受表明，这种关系让你感到不够安全和稳定。你提到过孩子，你觉得孩子对你的感情生活有怎样的影响？

Therapist: What you're describing shows that the relationship doesn't make you feel entirely safe or stable. You mentioned your child before—how do you think your child influences your romantic decisions?

来访者：孩子让我更加珍惜家庭的完整，我也害怕如果我在这段关系中受伤害，会影响孩子。

Client: Having a child makes me cherish the idea of a complete family more. I'm also afraid that if I get hurt in this relationship, it will affect my child.

咨询师：孩子的存在确实会在很大程度上影响你的选择。你提到孩子让你更加谨慎，那你觉得如何平衡自己的情感需求和孩子的利益？

Therapist: A child's presence definitely plays a significant role in the choices you make. Since you mentioned being more cautious because of your child, how do you think you can balance your emotional needs with their well-being?

来访者：我觉得我需要找到一个能够让我安心的人，这样我才能更好地照顾孩子。

Client: I think I need to find someone who makes me feel truly at ease, so I can take care of my child better.

咨询师：你希望找到一个能让你安心的人，这是一种合理的需求。你能描述一下，什么样的人或关系会让你感到安心吗？

Therapist: Wanting to be with someone who gives you peace of mind is a completely valid need. Can you describe what kind of person or relationship would make you feel secure?

来访者：我希望对方是一个值得信赖的人，他能给我足够的安全感和爱，让我知道我们之间是稳定的。

Client: I want someone trustworthy, who can give me enough love and a strong sense of security, so I know that our relationship is stable.

咨询师：你对安全感和爱的渴望是值得被关注的。你认为目前的关系中有多少部分能让你感到安全？

Therapist: Your desire for love and security is absolutely worth paying attention to. In your current relationship, how much of that security do you feel is actually present?

来访者：很少，我觉得他虽然对我很好，但他的态度让我始终不确定，这让我很难安心。

Client: Very little. He treats me well, but his unclear attitude always leaves me unsure. It's hard to feel at ease.

咨询师：你的感受是完全可以理解的。当一个人无法提供明确的承诺时，确实会让人感到不安。你觉得你目前能为自己做些什么来增加自己的安全感？

Therapist: Your feelings are completely valid. When someone can't provide a clear commitment, it naturally leads to insecurity. What do you think you can do right now to increase your sense of safety?

来访者：我可能需要更多时间来了解他，看看他是否真的能给我这份安全感。

Client: Maybe I need more time to get to know him and see whether he can really give me that sense of security.

咨询师：给自己更多时间去了解对方是个好主意。在这个过程中，你也可以更深入地了解自己的需求和底线，这会帮助你在未来的关系中做出更明智的选择。

Therapist: Giving yourself more time to understand him is a good idea. And during that process, you can also explore

your own needs and boundaries more clearly—that will help you make wiser decisions in future relationships.

来访者：我担心的是时间可能不够，如果最后发现他还是不能满足我的需求，我会更痛苦。

Client: What I'm worried about is, what if time runs out? What if I eventually find out he still can't meet my needs, and I end up feeling even worse?

咨询师：你的担忧是完全可以理解的。在这种情况下，你需要相信自己的直觉，同时也要学会在关系中设立健康的边界，以保护自己。

Therapist: That's a completely understandable concern. In this situation, it's important to trust your intuition while also learning to set healthy boundaries in the relationship—to protect yourself.

来访者：我知道我需要学会保护自己，但我还是忍不住想要靠近他。

Client: I know I need to protect myself, but I can't help wanting to be close to him.

咨询师：靠近他人是人之常情，尤其是在你感到孤独或受伤的时候。但同时，你也需要学会倾听自己的内心声音，明白什么时候该靠近，什么时候该保持距离。

Therapist: It's completely human to want closeness, especially when you're feeling lonely or vulnerable. At the same time, learning to listen to your inner voice can help you recognize when it's right to lean in, and when to step back.

来访者：你说得对，我需要学会倾听自己的内心。但我还是不知道该怎么做。

Client: You're right. I need to learn to listen to my inner voice. But I still don't know how.

咨询师：倾听自己的内心并不容易，但这是个重要的练习。你可以从每天花一点时间独处开始，问问自己真正想要的是什么，然后慢慢学会识别自己的情感和需求。

Therapist: Listening to your inner self isn't easy, but it's a very important practice. You can start by spending a bit of quiet time with yourself each day. Ask yourself what you truly want, and gradually learn to recognize your emotions and needs.

来访者：我明白了，我会试着这样做。谢谢你，咨询师，我感觉好多了。

Client: I understand now. I'll try to do that. Thank you, therapist. I feel much better.

咨询师：很高兴听到你感觉好多了。记住，这是一个过程，你不需要马上找到答案。我们下次见面时可以继续探讨，好吗？

Therapist: I'm really glad to hear that you're feeling better. Remember, this is a process—you don't have to figure it all out at once. We can continue exploring this next time, okay?

来访者：好的，谢谢你，我真的很感激你的帮助。

Client: Okay, thank you. I really appreciate your help.

咨询师：不客气，我会一直在这里支持你。下次见。

Therapist: You're very welcome. I'll be here to support you. See you next time.

来访者：下次见。

Client: See you next time.

咨询师：下次见。

Therapist: See you next time.

# PsyLLM-Large-250519

来访者：您好，咨询师老师。我最近感觉特别累，心里很乱，不知道该怎么办。

Client: Hello, Teacher Therapist. I've been feeling very tired lately, and my mind is in chaos. I don't know what to do.

咨询师：小雨，听到你说累和乱，我能感受到你现在内心的沉重和无助。你能告诉我，这种乱具体是什么样的感觉吗？是像一团纠结的线，还是有什么特别鲜明的情绪？

Therapist: Xiaoyu, hearing you say you're tired and confused, I can feel the heaviness and helplessness in your heart. Could you tell me what this chaos feels like specifically? Is it like a tangled ball of thread, or are there any particularly vivid emotions?

来访者：嗯……像一团乱线吧。一边是学业上的压力，觉得自己必须考上好大学，不然辜负了家人的期望；另一边是感情上的纠结，我和一个同学谈恋爱了，但家里完全不同意，他们觉得会影响我学习。

Client: Hmm... it's like a tangled ball of thread. On one side, there's academic pressure, feeling that I must get into a good university, otherwise I'll disappoint my family's expectations; on the other side, there's emotional entanglement - I'm in a relationship with a classmate, but my family completely disagrees, they think it will affect my studies.

咨询师：你内心似乎被两种力量拉扯着———一方面是对未来和家人的责任感和压力，另一方面是对这段感情的渴望和被认可的期待。这两种声音在你心里同时响起，让你感到难以喘息。这种复杂的情绪，能不能再帮我描述一下，它对你意味着什么？

Therapist: It seems your heart is being pulled in two directions - on one hand, there's the sense of responsibility and pressure for the future and family, and on the other, there's the desire for this relationship and the expectation of being accepted. These two voices are speaking simultaneously in your heart, making it hard to breathe. Could you help me understand more about these complex emotions, what do they mean to you?

来访者：有时候我觉得自己像个夹缝里的人，两边都不讨好。家里人觉得我变坏了，不像以前那样专注于学习；我自己又怕这段感情真的让我走偏，考不上理想的学校。那种害怕和自责交织在一起，让我很难受。

Client: Sometimes I feel like I'm stuck in a crack, not pleasing either side. My family thinks I've changed for the worse, not focusing on studies like before; I'm also afraid this relationship might really lead me astray, preventing me from getting into my ideal school. That fear and self-blame are intertwined, making me feel terrible.

咨询师：这夹缝中的感受非常真实，也充满了孤独和无助。你提到害怕和自责，这两种情绪往往让人陷入一种循环，觉得自己既不够好也不被理解。你觉得这种循环是从什么时候开始的？有没有特别明显的时刻或者记忆，让你感受到这种被撕裂的感觉？

Therapist: This feeling of being in a crack is very real and full of loneliness and helplessness. You mentioned fear and self-blame, these two emotions often trap people in a cycle, making them feel neither good enough nor understood. When do you think this cycle began? Are there any particular moments or memories when you felt this tearing sensation?

来访者：大概是恋爱开始后，尤其是看到成绩下降的时候。那一刻，我觉得自己失败了，好像辜负了所有人的期待。特别是妈妈，她很失望，甚至说我不懂事，这让我心里更难受了。

Client: It probably started after the relationship began, especially when I saw my grades dropping. At that moment, I felt like I had failed, as if I had disappointed everyone's expectations. Especially my mother, she was very disappointed, even saying I was being immature, which made me feel worse.

咨询师：妈妈的失望和批评像是放大镜，把你的内心挣扎放大了。那种被最亲近的人误解和否定，会让人感到深深的无力和羞愧。你愿意跟我分享，当你听到她的话时，身体上或心里有什么反应吗？比如紧张、难过，还是别的什么？

Therapist: Your mother's disappointment and criticism are like a magnifying glass, amplifying your inner struggle. That kind of misunderstanding and rejection from someone so close can make one feel deeply powerless and ashamed. Would you be willing to share what physical or emotional reactions you had when you heard her words? Like tension, sadness, or something else?

来访者：我会心跳加速，胸口发闷，有时候想躲起来，不想面对任何人。那种感觉像是有块石头压着我，让我喘不过气。

Client: My heart would race, my chest would feel tight, sometimes I want to hide, not wanting to face anyone. That feeling is like having a stone pressing on me, making it hard to breathe.

咨询师：谢谢你如此坦诚地分享这些身体和情绪的反应。心跳加速和胸闷都是身体在告诉你，你正在经历强烈的压力和恐惧。它们像一种警报，提醒我们需要关注和保护自己。你觉得，这样的警报声在你生活中还有哪些时刻会响起？或者，它是否和过去的某些经历有联系？

Therapist: Thank you for sharing these physical and emotional reactions so honestly. The racing heart and chest tightness are your body's way of telling you that you're experiencing intense pressure and fear. They're like an alarm, reminding us to pay attention to and protect ourselves. When else in your life do you think this alarm might sound? Or does it connect to any past experiences?

来访者：我想，可能是小时候考试没考好，爸爸也会批评我，那时我也会觉得胸口紧绷，好像自己永远达不到他们的要求。

Client: I think, maybe when I was young and didn't do well on exams, my father would criticize me, and I would feel that same chest tightness, as if I could never meet their expectations.

咨询师：这很重要，小雨。你刚才提到的童年经历，似乎在你内心埋下了一种对失败的恐惧和对不被爱的担忧。现在的学业压力和感情困境，可能不仅仅是当下的挑战，而是触发了那些深层的情感模式。你愿意和我一起慢慢探索这些旧有的感受吗？我们可以一步步来，不急。

Therapist: This is important, Xiaoyu. The childhood experience you just mentioned seems to have planted a fear of failure and worry about not being loved in your heart. The current academic pressure and relationship difficulties might not just be present challenges, but are triggering those deeper emotional patterns. Would you be willing to explore these old feelings together with me? We can take it step by step, no rush.

来访者：我愿意试试，只是有时候觉得很难，像是在揭开伤口。

Client: I'm willing to try, but sometimes it feels difficult, like opening up old wounds.

咨询师：这份觉察本身就很勇敢，也显示了你对自我成长的渴望。我们不急于解决所有问题，而是先让这些伤口被看见和理解。现在，回到你和同学的关系，你刚刚说这段感情带来了快乐，但也有压力和困惑。能否请你描述一下，当你们在一起时，最让你感到温暖或安全的瞬间是什么？

Therapist: This awareness itself is very brave and shows your desire for self-growth. We're not in a hurry to solve all problems, but first to let these wounds be seen and understood. Now, returning to your relationship with your classmate, you just said this relationship brings happiness but also pressure and confusion. Could you describe what moment with her makes you feel the most warm or safe?

来访者：有时候，只是和她并肩坐在教室外，聊一些无关紧要的小事，我就觉得心里很安稳，像是找到了一个可以依靠的港湾。那一刻，所有的烦恼都暂时消失了。

Client: Sometimes, just sitting next to her outside the classroom, talking about trivial things, I feel very peaceful, like finding a harbor I can rely on. In that moment, all worries temporarily disappear.

咨询师：这份安稳和依靠，是你内心深处对连接和归属感的真切渴望。它不仅仅是一种情感上的满足，更是一种心灵的慰藉。与此同时，家庭和学业带来的压力像是一堵墙，让你在这份渴望和安全之间感到矛盾和挣扎。这种矛盾，你觉得它对你的自我认知产生了什么样的影响？

Therapist: This peace and reliance is a genuine desire for connection and belonging in your heart. It's not just an emotional satisfaction, but also a spiritual comfort. Meanwhile, the pressure from family and academics is like a wall, making you feel conflicted and struggling between this desire and safety. How do you think this conflict has affected your self-perception?

来访者：我觉得自己变得很分裂。一面是那个努力想成为家人眼中优秀孩子的自己，另一面是想要自由和被理解的自己。这两者好像在打架，我不知道该听谁的。

Client: I feel like I've become very split. One side is the self that tries hard to be the excellent child in my family's eyes, the other side is the self that wants freedom and understanding. These two seem to be fighting, I don't know which one to listen to.

咨询师：这种内在的冲突确实会让人感到疲惫和迷失。你用打架这个词，生动地表达了内心的拉扯。其实，这种分裂并不代表你有错，而是成长中一个重要的课题。你已经开始意识到这两个部分的自己，这本身就是一种深刻的觉察和勇气。你愿意和我一起，慢慢探寻如何在尊重这两个不同的声音的同时，找到属于你自己的平衡点吗？

Therapist: This inner conflict can indeed make one feel exhausted and lost. Your use of the word "fighting" vividly expresses the inner tug-of-war. Actually, this split doesn't mean you're wrong, but rather an important lesson in growth. You've already begun to recognize these two parts of yourself, which in itself is a profound awareness and courage. Would you be willing to explore together how to find your own balance point while respecting these two different voices?

来访者：我想找那个平衡点，可是感觉好难，尤其是当家里的压力那么大的时候。

Client: I want to find that balance point, but it feels so hard, especially when the family pressure is so great.

咨询师：我听到你对找到平衡点的渴望，也感受到这份渴望背后深深的疲惫和无助。面对家人的强烈反对，那种无力和挫败感一定很重。你提到过，你曾经试图跟妈妈沟通，但结果并不顺利。能否分享一下那次对话中，你最深的感受是什么？

Therapist: I hear your desire to find a balance point, and I also feel the deep exhaustion and helplessness behind this desire. Facing strong family opposition, that sense of powerlessness and frustration must be heavy. You mentioned that you once tried to communicate with your mother, but it didn't go well. Could you share what your deepest feeling was during that conversation?

来访者：我觉得自己被误解了，她只看到了成绩的下降，没有看到我内心的挣扎和痛苦。我说话的时候，她常常打断我，甚至语气变得严厉，最后我只能选择沉默，因为再怎么说也没用。

Client: I felt misunderstood, she only saw the drop in grades, not my inner struggle and pain. When I was speaking, she often interrupted me, even becoming harsh in tone, and in the end I could only choose silence, because there was no use

in saying more.

咨询师：在那样的沟通氛围中，你感受到被忽视和孤立，这种无力感像是一道无形的屏障，阻碍了你表达真实感受的通道。你的沉默，也许是一种保护自己不再受伤的方式。你愿意和我一起，探讨如何找到一种既能保护你自己，又能尝试重新打开对话的方法吗？我们可以慢慢来，不急于一时。

Therapist: In that kind of communication atmosphere, you felt ignored and isolated, this sense of powerlessness was like an invisible barrier, blocking your channel to express your true feelings. Your silence might be a way to protect yourself from further hurt. Would you be willing to explore together how to find a method that can both protect yourself and try to reopen the dialogue? We can take it slowly, no rush.

来访者：我想试，但我怕结果还是会让我失望，怕再次被拒绝。

Client: I want to try, but I'm afraid the result will still disappoint me, afraid of being rejected again.

咨询师：这份害怕非常真实，也很合理。你经历了多次努力却未获回应，心里的防御机制自然启动，试图保护你不被再次伤害。我们不必急着去突破这个防线，而是可以先在这里，给予这份害怕空间，允许它存在。你愿意和我一起，先温柔地陪伴这份害怕，听它想告诉你什么吗？

Therapist: This fear is very real and reasonable. You've experienced multiple attempts without response, so your psychological defense mechanism naturally activates to protect you from being hurt again. We don't need to rush to break through this defense, but can first give this fear space, allow it to exist. Would you be willing to gently accompany this fear with me, listen to what it wants to tell you?

来访者：嗯，我想让它被听见，也许它一直在提醒我别再受伤。

Client: Yes, I want it to be heard, maybe it's been reminding me not to get hurt again.

咨询师：这是一份宝贵的内心智慧。你的害怕在提醒你，关系中的脆弱和不确定性。它也是你内心对安全感的深刻渴望。接下来，我们可以一起探索如何在尊重这份害怕的基础上，逐步建立一种更有力量的沟通方式，不仅帮你表达自己，也帮助你守护自己的边界。

Therapist: This is precious inner wisdom. Your fear is reminding you of the vulnerability and uncertainty in relationships. It's also your heart's deep desire for security. Next, we can explore together how to gradually build a more powerful way of communication based on respecting this fear, not only helping you express yourself but also helping you protect your boundaries.

来访者：听您这么说，我感觉好像有了些方向，不那么孤单了。

Client: Hearing you say this, I feel like I have some direction, not so alone anymore.

咨询师：这就是我们的起点——在一个安全的空间里，让那些复杂的情绪被看见和理解。你今天的分享，已经让我们触及了你内心最柔软的部分。我们可以继续沿着这条线索，慢慢地探索你对自己的理解，以及对未来的期待。你觉得，这样的节奏对你来说合适吗？

Therapist: This is our starting point - in a safe space, letting those complex emotions be seen and understood. Your sharing today has already allowed us to touch the softest parts of your heart. We can continue along this thread, slowly exploring your understanding of yourself and your expectations for the future. Do you think this pace is suitable for you?

来访者：是的，我觉得这样挺好，不急，也不放弃自己。

Client: Yes, I think this is good, not rushing, and not giving up on myself.

咨询师：很好，小雨。你的这份耐心和对自己温柔的对待，是内心强大的体现。接下来，我们可以一起回顾一下今天的内容，看看哪些部分对你触动最深，也可以为你接下来的行动带来启发。你觉得呢？

Therapist: Very good, Xiaoyu. This patience and gentle treatment of yourself is a manifestation of inner strength. Next, we can review today's content together, see which parts touched you the most, and also bring inspiration for your next actions. What do you think?

来访者：好的，我想回顾一下。

Client: Okay, I'd like to review.

咨询师：谢谢你的配合。今天我们谈到了你在学业和亲密关系中的双重压力，家庭的期望和批评引发的内心冲突，以及童年经历的影子。你也分享了那种被撕裂、孤独和无助的感受，同时也发现了内心渴望被理解和安全感的需求。你还表达了尝试沟通的勇气和面对害怕的脆弱。这些都是非常重要的发现。你愿意告诉我，哪一个部分对你来说最触动，或者最想带回去思考和实践的吗？

Therapist: Thank you for your cooperation. Today we discussed the dual pressure you face in academics and intimate relationships, the inner conflict triggered by family expectations and criticism, and the shadow of childhood experiences. You also shared the feeling of being torn, lonely and helpless, while discovering your inner need for understanding and security. You also expressed the courage to try communication and the vulnerability in facing fear. These are all very important discoveries. Would you tell me which part touched you the most, or which you most want to take back to think about and practice?

来访者：我觉得最触动的是那个夹缝里的感受，还有害怕被拒绝时的无助。我想试着用更温柔的方式面对自己，不让那个害怕的声音淹没我。

Client: I think what touched me most was the feeling of being in a crack, and the helplessness when afraid of rejection. I want to try to face myself more gently, not letting that fearful voice overwhelm me.

咨询师：这是一个非常有意义的觉察。面对内心的恐惧和伤痛，给自己温柔和接纳，是我们走向成长的重要一步。我们可以把这作为一个小小的起点，在接下来的时间里，继续关注和练习如何与这些情绪共处，而不是对抗它们。你愿意在日常生活中，给自己留出一些时间，去感受和聆听这些情绪吗？哪怕只是短暂的片刻。

Therapist: This is a very meaningful awareness. Facing inner fears and wounds, giving yourself gentleness and acceptance is an important step toward growth. We can take this as a small starting point, and in the coming time, continue to focus on and practice how to coexist with these emotions, rather than fighting against them. Would you be willing to set aside some time in your daily life to feel and listen to these emotions? Even if it's just for a brief moment.

来访者：愿意，我想试试看。

Client: Yes, I want to try.



\end{document}